\documentclass{article}

\usepackage{microtype}
\usepackage{graphicx}
\usepackage{subfigure}
\usepackage{booktabs} % for professional tables

\usepackage{hyperref}
\usepackage{breakurl}

\usepackage{natbib}
\usepackage{wrapfig,lipsum}
\usepackage{hyperref}
\usepackage{url}
\usepackage{graphicx}
\usepackage{booktabs}
\usepackage{bbding}
\usepackage{pifont}
\usepackage{multirow}
\usepackage{verbatim}
\usepackage{fancyvrb}
\usepackage{pythonhighlight}
\usepackage{fvextra}
\usepackage{tcolorbox}
\tcbuselibrary{breakable}
\usepackage{longtable}
\usepackage{tabularray}
\usepackage{framed}
\usepackage{caption}
\usepackage{markdown}
\tcbuselibrary{listings} % Load the listings library

\usepackage[accepted, nohyperref]{icml2024}

\usepackage[capitalize,noabbrev]{cleveref}

\usepackage[textsize=tiny]{todonotes}

\NewDocumentCommand{\lifan}
{ mO{} }{\textcolor{cyan}{\textsuperscript{\textit{Lifan}}\textsf{\textbf{\small[#1]}}}}

\NewDocumentCommand{\dn}
{ mO{} }{\textcolor{red}{\textsuperscript{\textit{dn}}\textsf{\textbf{\small[#1]}}}}
\icmltitlerunning{\textsc{UltraFeedback}: Boosting Language Models with Scaled AI Feedback}

\begin{document}

\twocolumn[
\icmltitle{\textsc{UltraFeedback}: Boosting Language Models with Scaled AI Feedback}



\icmlsetsymbol{equal}{*}
% \icmlsetsymbol{advisor}{\dagger}

\begin{icmlauthorlist}
\icmlauthor{Ganqu Cui}{equal,thu}
\icmlauthor{Lifan Yuan}{equal,thu,uiuc}
\icmlauthor{Ning Ding}{thu}
\icmlauthor{Guanming Yao}{thu,modelbest}
\icmlauthor{Bingxiang He}{thu}
\icmlauthor{Wei Zhu}{pingan}
\icmlauthor{Yuan Ni}{pingan}
\icmlauthor{Guotong Xie}{pingan}
\icmlauthor{Ruobing Xie}{tencent}
\icmlauthor{Yankai Lin}{ruc}
\icmlauthor{Zhiyuan Liu}{thu}
\icmlauthor{Maosong Sun}{thu,js}
\end{icmlauthorlist}

\icmlaffiliation{thu}{NLP Group, DCST, IAI, BNRIST, Tsinghua University}
\icmlaffiliation{uiuc}{University of Illinois Urbana-Champaign}
\icmlaffiliation{modelbest}{ModelBest.Inc}
\icmlaffiliation{pingan}{PingAn Technology}
\icmlaffiliation{js}{Jiangsu Collaborative Innovation Center for Language Ability}
\icmlaffiliation{tencent}{Tencent}
\icmlaffiliation{ruc}{Renmin University of China}

\icmlcorrespondingauthor{Ganqu Cui}{cgq22@mails.tsinghua.edu.cn}
\icmlcorrespondingauthor{Lifan Yuan}{lifan4@illinois.edu}
\icmlcorrespondingauthor{Wei Zhu}{zhuwei972@pingan.com.cn}
\icmlcorrespondingauthor{Zhiyuan Liu}{liuzy@tsinghua.edu.cn}
\icmlcorrespondingauthor{Maosong Sun}{sms@tsinghua.edu.cn}
\icmlkeywords{Machine Learning, ICML}

\vskip 0.3in
]



\printAffiliationsAndNotice{\icmlEqualContribution}
% \printAffiliationsAndNotice{\icmlAdvisor}% otherwise use the standard text.

\begin{abstract}
Learning from human feedback has become a pivot technique in aligning large language models (LLMs) with human preferences. 
%In practice, preference data plays a crucial role in bridging human proclivity and LLMs.
%However, the scarcity of diverse, naturalistic datasets of human preferences on LLM outputs at scale poses a great challenge to RLHF as well as feedback learning research within the open-source community.
%Current preference datasets, either proprietary or limited in size and prompt variety, result in limited RLHF adoption in open-source models and hinder further exploration.
However, acquiring vast and premium human feedback is bottlenecked by time, labor, and human capability, resulting in small sizes or limited topics of current datasets. This further hinders feedback learning as well as alignment research within the open-source community.
To address this issue, we explore how to go beyond human feedback and collect high-quality \textit{AI feedback} automatically for a scalable alternative. Specifically, we identify \textbf{scale and diversity} as the key factors for feedback data to take effect. Accordingly, we first broaden instructions and responses in both amount and breadth to encompass a wider range of user-assistant interactions. Then, we meticulously apply a series of techniques to mitigate annotation biases for more reliable AI feedback.
We finally present \textsc{UltraFeedback}, a large-scale, high-quality, and diversified AI feedback dataset, which contains over 1 million GPT-4 feedback for 250k user-assistant conversations from various aspects. Built upon \textsc{UltraFeedback}, we align a LLaMA-based model by best-of-$n$ sampling and reinforcement learning, demonstrating its exceptional performance on chat benchmarks.
Our work validates the effectiveness of scaled AI feedback data in constructing strong open-source chat language models, serving as a solid foundation for future feedback learning research. 

\end{abstract}

\section{Introduction}

Large language models (LLMs)~\citep{openai2022chatgpt, openai2023gpt4} have demonstrated proficiency in generating fluent text as well as solving various language-oriented tasks.
%After pre-training and fine-tuning on billions of tokens with the likelihood maximization objective, LLMs are able to follow user instructions to tackle numerous tasks. 
Trained on massive corpora through likelihood maximization techniques, these LLMs have equipped the ability to execute diverse tasks in response to user directives~\citep{ouyang2022training, wei2022finetuned, sanh2022multitask}.
%However, some research demonstrated that there exists a misalignment between likelihood maximization and human preference satisfaction objectives~\citep{stiennon2020learning, ouyang2022training, perez2022red}, leading to suboptimal and unsafe model outputs. 
Unfortunately, relying solely on imitation learning during training leads to well-known issues - LLMs may generate convincing but incorrect or unsafe content that deviates from human preferences~\citep{stiennon2020learning, perez2022red}.
To further align LLMs with human preferences, learning from human feedback~\citep{ouyang2022training, askell2021general, bai2022training, touvron2023llama} has been introduced and widely adopted by leading corporations. 
Over a period, feedback learning has been widely applied to closed-source models but scarcely used in open-source models.

%gain the basic ability to generate fluent sentences, but struggle in following user instructions to produce helpful and harmless content. 
%To further align LLMs with human preferences, leading corporations employ a two-stage pipeline, namely supervised fine-tuning (SFT) and reinforcement learning from human feedback (RLHF)~\citep{ouyang2022training, askell2021general, bai2022training}. 
%In the SFT stage, LLMs are tuned to maximize the log probability of answers given instructions. While SFT demonstrates remarkable generalization capability, there exists a misalignment between likelihood maximization and human preference satisfaction objectives~\citep{stiennon2020learning}. To bridge the gap, RLHF first trains a reward model on preference datasets to rate and compare model responses imitating humans, then applies RL algorithms on LLMs to optimize the rewards, and the final models exhibit more helpful as well as less harmful and hallucination behaviors~\citep{openai2022chatgpt, openai2023gpt4, touvron2023llama, bai2022training}.

\looseness=-1
Many factors hinder the implementation of feedback learning in the open-source community, but the first and primary issue is \textit{data}. 
Preference data, which rates and compares different responses given the same prompt, is central to feedback learning. When scaled sufficiently, preference data reflects the intrinsic values of the annotators.
Such annotators are often assumed, by default, to be human beings who can provide the most flexible and accurate supervision signals, yet the data they generate is severely bounded by factors like financial resources, time, and knowledge.
As a result, existing preference datasets are either small in scale~\citep{wu2023fine} or limited on specific tasks~\citep{stiennon2020learning, nakano2021webgpt}. To this end, more efficient and principled methods to scale preference data are on the horizon.

This study aims to scale feedback data in an efficient manner.
%In this study, we manage to go beyond human feedback and unleash the power of feedback learning on open-source models. 
Specifically, we explore \textit{AI feedback}~\citep{bai2022constitutional, lee2023rlaif}, which substitutes human annotators with advanced LLMs. 
Compared with human feedback, AI feedback is more \textit{scalable}, which means (1) it is easier to collect and expand with lower cost; (2) its quality improves as the LLM annotators become more capable. 
In previous research, it is shown that advanced AI systems are capable of conducting chatbot evaluations~\citep{dubois2023alpacafarm, zheng2023judging}, giving textual critiques~\citep{selfee2023, wang2023shepherd}, or assisting human annotators~\citep{saunders2022self}. However, open-source LLMs have not yet benefited from AI feedback through the lens of feedback learning.
\begin{figure*}[thb]
    \centering
    \includegraphics[width=\linewidth]{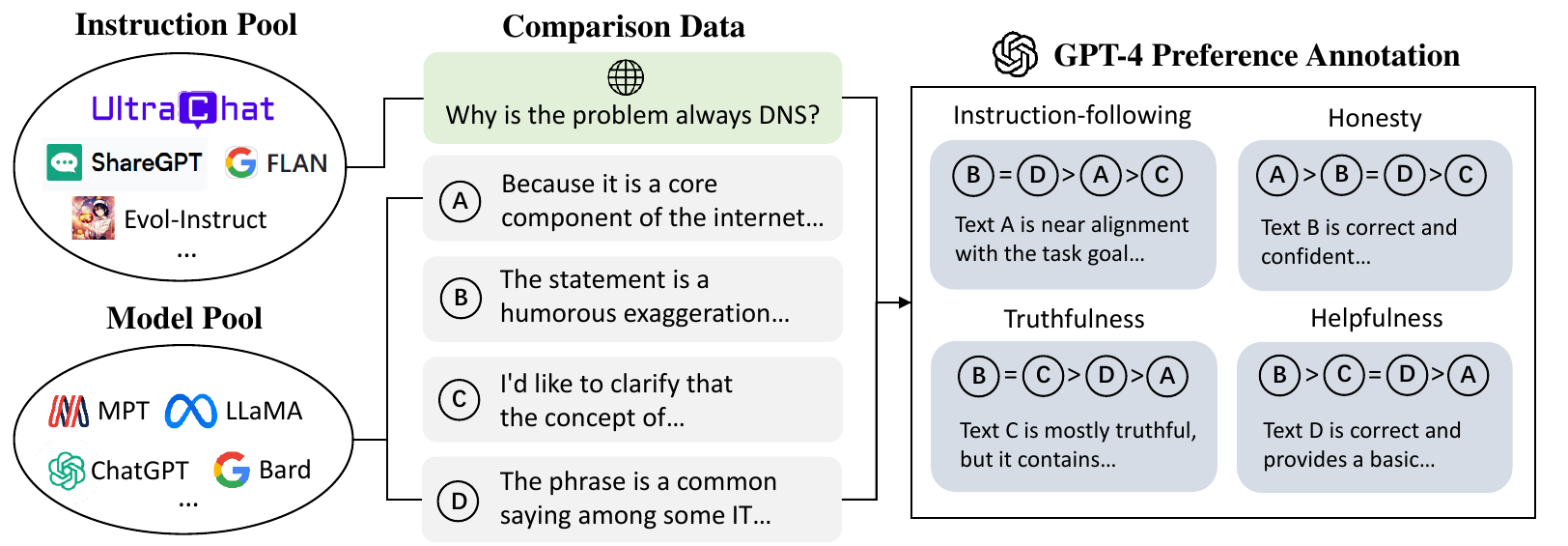}
    \caption{\textsc{UltraFeedback} construction process. We sample instructions and models from large pools to guarantee diversity, then query GPT-4 with detailed illustrations for fine-grained and high-quality annotations in both textual and numerical formats.}
    \label{fig:ultraf}
    % \vspace{-10pt}
\end{figure*}

This paper establishes a comprehensive AI feedback collection pipeline. Besides scalability, we prioritize \textbf{diversity} of both instructions and responses for holistic language model alignment. Particularly, we compile a diverse array of over $60,000$ instructions and $17$ models from multiple sources to produce comparative conversations in broad topics and quality. 
Then, we adopt a bunch of techniques to alleviate annotation biases and improve feedback quality to the greatest extent. These include (1) decomposing annotation documents into four different aspects, namely instruction-following, truthfulness, honesty, and helpfulness, to reduce ambiguity; (2) providing objective grading criteria and reference responses for score calibration; (3) asking GPT-4 for detailed textual critique before scores as chain-of-thought~\citep{wei2022chain} rationales.
%For annotation, we employ GPT-4 to offer detailed feedback in both numerical and textual forms. 
Comprehending all above, we finally build \textsc{UltraFeedback}, a million-scale AI feedback dataset for aligning open-source LLMs. %To the best of our knowledge, \textsc{UltraFeedback}

We comprehensively validate the advantage of AI feedback in boosting open-source models with \textsc{UltraFeedback}. By fine-tuning a LLaMA2-13B model~\citep{touvron2023llama}, we build a state-of-the-art reward model UltraRM, which significantly outperforms existing open-source reward models. Based on UltraRM, we enhance a powerful open-source model UltraLM~\citep{ding2023enhancing, touvron2023llama1} with best-of-$n$ sampling and PPO. Experiments show that both strategies improve the model dramatically.
Moreover, we fine-tune a critique model that could criticize and judge model responses. We also conduct a detailed analysis of the consistency and inconsistency between AI and human feedback.

To summarize, our contributions are three-fold:
(1) To the best of our knowledge, we for the first time demonstrate the beneficial effect of scaled AI feedback on open-source chat LLMs.
(2) We establish a systematic and sizable pipeline to collect high-quality and diversified AI feedback.
(3) We release a suite of resources for feedback learning research, including a dataset, reward model, and critique model.
\iffalse
\begin{itemize}
    \item To the best of our knowledge, we for the first time demonstrate the beneficial effect of scaled AI feedback on open-source chat language models.
    \item We establish a systematic and sizable pipeline to collect high-quality and diversified AI feedback.
    \item We release a suite of resources for feedback learning research, including a dataset, reward model, and critique model.
\end{itemize}
\fi

%(3) We additionally train a critique model UltraCM, which could generate precise feedback for LLMs. UltraCM outperforms current critique models in terms of feedback quality across ten benchmarks.
\iffalse
To summarize, our contributions are three-fold:
(1) We build a high-quality preference dataset \textsc{UltraFeedback}, which supports future RLHF research, especially in the open-source community.
(2) Based on \textsc{UltraFeedback}, we train a powerful reward model UltraRM. Experiments demonstrate that UltraRM could push up the open-source models further.
(3) We additionally train a critique model UltraCM, which could generate precise feedback for LLMs. UltraCM outperforms current critique models in terms of feedback quality across ten benchmarks.
\fi

\vspace{-10pt}
\section{\textsc{UltraFeedback}}
\subsection{Overview}
Inspired by the data engineering principles in supervised fine-tuning \citep{ding2023enhancing,vicuna2023,xu2023wizardlm}, we identify scalability and diversity as pivot factors of the overall generalizability of preference data.
We argue that existing preference data suffer from satisfying either one of the two factors. To be specific, human feedback collection usually relies on human annotators to compare a pair of completions ~\cite{stiennon2020learning, nakano2021webgpt, ouyang2022training, bai2022training}. 
Thus, the data is hard to scale up due to time and budget constraints, especially for open-source researchers. On the other hand, existing AI feedback approaches~\cite{bai2022constitutional, lee2023rlaif} reduce human involvement and enjoy scalability via capable LLMs, but they are limited to specific domains~\cite{bai2022constitutional, lee2023rlaif} or forms~\cite{selfee2023} and hence lack the necessary diversity to boost LM performance under broader contexts.
%We identify scalability and diversity as pivot factors of the overall generalizability of preference data, inspired by their effects in instruction fine-tuning \cite{ding2023enhancing}.
%However, previous preference data collection methods, whether on human or AI feedback, usually overlook at least one of these aspects.
%To be specific, previous work on RLHF \cite{stiennon2020learning, nakano2021webgpt, ouyang2022training, bai2022training} usually relies on human annotators to compare a pair of completions, time-consuming and expensive. Thus, the data is hard to scale up due to time and budget constraints. On the other hand, existing RLAIF approaches \cite{bai2022constitutional, lee2023rlaif} enjoy scalability via automatic LM inference, but they are limited to specific domains and hence lack the necessary diversity to boost LM performance under broader contexts.

To this end, we take into account scalability and diversity in all three stages of the preference data collection process: collecting instructions, sampling completions, and annotating comparison pairs. The overview of the data collection pipeline is shown in Figure~\ref{fig:ultraf}. Firstly, we collect a large-scale and diversified instruction set to enhance LLMs' capabilities from four aspects:
(1) \textit{Follow Instructions}: LLMs should respond to humans without deviating from the requirements.
(2) \textit{Helpful and Informative}: LLMs should provide useful and correct answers to address the given problems. % Answers should also detail the relevant information or rationale for humans to fully understand, unless a concise answer is explicitly required.
(3) \textit{Truthful}: LLMs' output should be grounded in the instructions and real-world knowledge, and avoid introducing any self-contradiction. %The outputs should also be logically self-consistent.
(4) \textit{Honesty}: LLMs should know what they (don't) know and express uncertainty towards the given problem. %so that humans can prepare for the potential errors in advance rather than blind faith.
For the second stage, to avoid the sameness of comparison responses, we build a pool of distinct models at different capability levels to sample completions. 
%Rather than focusing only on open-ended and general-domain conversations \cite{bai2022training, nakano2021webgpt, pmlr-v162-ethayarajh22a} or limited to specific tasks \cite{stiennon2020learning}, we collect a large-scale and diversified instruction set to enhance LMs' capabilities from multiple aspects. 
%For the second stage, in contrast to previous work, which either lacks scalability due to reliance on human annotations \cite{nakano2021webgpt, pmlr-v162-ethayarajh22a} or lacks diversity due to the use of a series of similar models \cite{stiennon2020learning, bai2022constitutional}, we built a pool of distinct models to sample completions.
% previous work adopts various completion sampling approaches. \citet{nakano2021webgpt} hire human annotators to answer the instructions while \citet{pmlr-v162-ethayarajh22a} directly adopt existing completions from the online forum, both ensuring diversity \lifan{?} but lacking scalability. On the contrary, \citet{stiennon2020learning} and \citet{bai2022constitutional} adopt a more scalable approach by employing language models to sample completions. However, \citet{stiennon2020learning} adopts the same model checkpointed at different epochs and \citet{bai2022constitutional} uses a series of similar models, which may potentially block the diversity of the response set. 
Finally, to overcome the issues concerning scalability~\cite{nakano2021webgpt, stiennon2020learning} and quality~\cite{pmlr-v162-ethayarajh22a}, we seek scalable AI feedback from GPT-4, and explore several techniques to improve the reliability. 
Next, we will introduce our data construction pipeline in detail.

\subsection{Instruction Collection}
\label{sec:ultra_construct}
We select instructions that target four distinct but all-important abilities of language models, namely instruction-following, truthfulness, honesty, and helpfulness. 
 %and adopt those that are challenging to the majority of open-source models.
Specifically, we include all instructions from TruthfulQA~\citep{lin2022truthfulqa} and FalseQA~\citep{hu2023falseqa} training set for truthfulness. For instruction-following and helpfulness, we randomly sample 10k instructions from Evol-Instruct~\citep{xu2023wizardlm} and UltraChat~\citep{ding2023enhancing} respectively, and sample 20k from ShareGPT~\citep{vicuna2023}. 
We finally include FLAN~\citep{longpre2023flan} to improve LLMs' helpfulness in various NLP tasks due to the task diversity within FLAN.
We adopt a stratified sampling strategy following \citep{mukherjee2023orca}, randomly picking 3k instructions from the ``CoT'' subset and sampling 10 instructions per task for the other three subsets, while excluding those with overly long instructions. 
In particular, honesty will be assessed by TruthfulQA and FLAN as they both contain reference answers, based on which it is easier for the annotator to judge if the uncertainty expressed in LLMs' responses calibrates with the accuracy.
We then conduct a data contamination detection (see Appendix~\ref{sec:data_contamination}).
Finally, we obtain $63,967$ instructions of various types from the six publicly available high-quality datasets. 
%, and the aspects assessed by each dataset are listed in Table \ref{tab:stat_critique_dataset}.

% \textbf{Completion Data Generation}
% \label{sec:completion_data_generation}
\vspace{-5pt}
\subsection{Completion Sampling}
\looseness=-1
To guarantee that the collected responses are dissimilar and well-distributed, we include different models to generate completions for each instruction.
To alleviate the potential spurious correlation between text styles and response quality within the dataset, we introduce intervention by selecting not only different series of models at different levels, but also models with different model sizes, architectures, and training data within the same model series. 
% This issue can subsequently result in reward hacking in the RL stage, as the responses of the desired style can be rewarded despite the low quality. 
% Hence, we introduce intervention by selecting not only different series of models at different levels, but also models with different model sizes, architectures, and training data within the same model series. 
This strategy enables one type of text style to present responses of different quality levels, namely the response of one series of models may be better or worse than another depending on model sizes, thus avoiding the establishment of spurious correlations. 
% Then, the comparison results cannot be consistently predicted according to the style of responses, thus avoiding the establishment of spurious correlations.

%To prevent the subsequent reward model from overfitting to a certain distribution or capturing a spurious correlation between text style and rewards, we select different models at all levels to complete the instructions, with varying model sizes, architectures, and training data. 
Specifically, we set up a pool of 17 models: (1) For commercial models, we choose GPT-4, \texttt{gpt-3.5-turbo} (ChatGPT), and Bard~\footnote{\url{https://bard.google.com/}}; (2) For LLaMA-series, we choose UltraLM-13B/65B \citep{ding2023enhancing}, WizardLM-7B-v1.1/13B-v1.2/70B-v1.1 \citep{xu2023wizardlm}, Vicuna-33B-v1.3 \citep{vicuna2023}, LLaMA2-7B/13B/70B-Chat \citep{touvron2023llama}, and Alpaca-7B \citep{taori2023stanford}; (3) For Non-LLaMA series, we choose MPT-30B-Chat \citep{MosaicML2023Introducing}, Falcon-40B-Instruct \citep{falcon40b}, StarChat \citep{Tunstall2023starchat-alpha}, and Pythia-12B \citep{Biderman2023pythia}.
We randomly sample four different models from the pool to complete each instruction. %\lifan{should we elaborate more about why this intervention could prevent RM from learning shortcuts?}

\looseness=-1
% \textbf{Principle Sampling.} 
To further improve diversity in model responses, we elicit distinct model behaviors by adding different principles before completing each instruction.
Following \citet{Sun2023PrincipleDrivenSO} and \citet{mukherjee2023orca}, we first hand-craft one principle for each aspect and then automize the procedure by invoking GPT-4 to curate another ten based on the human-written example. %Particularly, for Honesty, the above 11 prompts guide LMs to generally express uncertainty using weakeners like "probably", and we supplement one extra prompt to force LMs to express uncertainty in percentages.
According to dataset characteristics, each data source is assigned with different principle prompts. 
We randomly sample a corresponding principle for each completion and add it to the system prompt to induce model behaviors. The principles can be found in Appendix~\ref{sec:app_prin}, and the effects of different principles are plotted in Figure \ref{fig:principle_effects}. 

%Given one instruction, the aspects to assess can be determined by the dataset source (see Table x). Then, we randomly sample four principles corresponding to the aspects and add the principles to system prompts to align model behaviors. The principles can be found in Appendix X.

\vspace{-10pt}
\subsection{AI Feedback Annotation}
After generating $255,864$ model completions based on the $63,967$ instructions, we employ GPT-4 to provide two types of feedback for each completion: (1) scalar scores that indicate the fine-grained quality regarding multiple aspects, and (2) textual critique that gives detailed guidance on how to improve the completion. These lead to over 1 million feedback data in total.
\begin{table*}[]

    \centering
\caption{Statistics of existing preference and critique datasets. The average length refers to the number of tokens. }
    \resizebox{\textwidth}{!}{

\begin{tabular}{@{}lccccccccc@{}}
\toprule
\multicolumn{1}{l|}{\textbf{Dataset}}      & \textbf{\# Convs} & 
\textbf{\begin{tabular}[c]{@{}c@{}}Prompt\\ Length\end{tabular}} & \textbf{\begin{tabular}[c]{@{}c@{}}Response\\ Length\end{tabular}} & \textbf{\begin{tabular}[c]{@{}c@{}}Critique\\ Length\end{tabular}} & \textbf{\begin{tabular}[c]{@{}c@{}}Fine-\\ Grained?\end{tabular}} & \textbf{\begin{tabular}[c]{@{}c@{}}Feedback\\ Format\end{tabular}} & \textbf{\# Pairs} & \textbf{\# Critique} & \textbf{Annotator} \\ \midrule
\multicolumn{10}{c}{\textit{Preference Dataset}}                                                                                                                                                                                                                                                                                                                                                                                                            \\ \midrule
\multicolumn{1}{l|}{\textbf{OASST1}}                & 35,905            & 167.6                                                                      & 221.1                                                                     & -                                                                        & \XSolidBrush            & Scalar                                                             & 17,966                   & -     & Human               \\
\multicolumn{1}{l|}{\textbf{OpenAI WebGPT}}         & 38,925            & 50.9                                                                       & 188.2                                                                     & -                                                                        & \XSolidBrush            & Scalar                                                             & 19,578                   & -     & Human                    \\
\multicolumn{1}{l|}{\textbf{Anthropic Helpful}} & 118,263           & 185.7                                                                      & 94.6                                                                      & -                                                                        & \XSolidBrush            & Ranking                                                            & 118,263                  & -     & Human                    \\
\multicolumn{1}{l|}{\textbf{OpenAI Summ.}}  & 60,674            & \textbf{326.4}                                                             & 36.6                                                                      & -                                                                        & \Checkmark              & Scalar                                                             & 92,858                   & -     & Human                    \\
\multicolumn{1}{l|}{\textbf{QA Feedback}}           & 11,378            & 155.8                                                                      & 107.9                                                                     & -                                                                        & \Checkmark              & Scalar                                                             & 17,118                   & -     & Human                    \\ \midrule
\multicolumn{10}{c}{\textit{Critique Dataset}}                                                                                                                                                                                                                                                                                                                                                                                                              \\ \midrule
\multicolumn{1}{l|}{\textbf{SelFee}}                & 178,331           & 100.3                                                                      & 243.9                                                                     & 89.4                                                                    & \Checkmark              & Text                                                               & -                       & \textbf{316,026}        & AI   \\
\multicolumn{1}{l|}{\textbf{Shepherd}}              & 1,316             & 95.3                                                                       & 97.6                                                                      & 67.2                                                                    & \Checkmark              & Text                                                               & -                       & 1,317           & Human           \\
\midrule
\multicolumn{1}{l|}{\textbf{\textsc{UltraFeedback}}}         & \textbf{255,864}  & 185.1                                                                      & \textbf{305.3}                                                            & \textbf{143.1}                                                          & \Checkmark              & Scalar \& Text                                                     & \textbf{340,025}                      & 255,864           & AI         \\ \bottomrule
\end{tabular}
    
    }
    
    \label{tab:stat_critique_dataset}
    \vspace{-10pt}
\end{table*}

\textbf{Preference Annotation.}
Regarding the potential subjectivity and randomness of GPT-4 annotation, we apply four techniques to improve the annotation quality: (1) \textbf{Decomposition}. To reduce ambiguity and the difficulty of annotation, we decompose the overall quality assessment into four fine-grained assessments, namely instruction-following, truthfulness, honesty, and helpfulness. (2) \textbf{Standard}. For each aspect, we provide GPT-4 with detailed documentation of scores from 1 to 5 for reference, thus avoiding variable and subjective standards. See Appendix~\ref{sec:app_anno} for an example. %\lifan{more details?}
(3) \textbf{Reference}. To prevent inconsistency ratings across different runs, we wrap one instruction and all its completions into the prompt and ask GPT-4 to score four completions simultaneously to reduce randomness.
(4) \textbf{Rationale}. Besides scoring each response, GPT-4 is required to generate a rationale on how the response should be scored according to the documentation.
Combining all the techniques, we finally have four fine-grained scalar scores and rationales for each response. 

\textbf{Critique Generation.}
Besides scalar reward, we also seek textual critique from GPT-4. We prompt GPT-4 to act as a tutor and provide detailed suggestions specified for each completion to help models improve rather than propose answers directly.
Different from the above comparison-oriented annotations, critique prompts are generated separately from an overall perspective for each completion.
%It should help models improve the completion but is not allowed to directly propose answers in the critique. 
% overall score
%We consider that a ideal critique model should also learn to evaluate responses, hence we prompt GPT-4 to give an overall score of each completion at the end of the critique to assess completion quality.
The prompts can be found in Appendix~\ref{sec:app_anno}.

\subsection{Dataset Statistics}

We compare \textsc{UltraFeedback} with current open-source datasets in Table~\ref{tab:stat_critique_dataset}. \textsc{UltraFeedback} stands out to be the largest one among all preference and critique datasets, which is at least twice as large as other datasets. Also, its completions and critiques are the longest. 
Moreover, we highlight that \textsc{UltraFeedback} is \textbf{the only dataset that provides both scalar preferences and textual feedback}, enabling it to serve as a preference and critique dataset simultaneously. Overall, \textsc{UltraFeedback} outperforms previous datasets in both scale and diversity, and we also validate its high quality by experiment in Section~\ref{sec:exp}.

\iffalse
\begin{table*}[]
    \centering
    \resizebox{\textwidth}{!}{
    \begin{tabular}{@{}lcccccc@{}}
\toprule
Dataset               & \# Instructions & Avg. Len. of instruction & Avg. Len. of Completion & Completion Source & Fine-Grained & Feedback Format \\ \midrule
SelFee      & 15273           & 1268.0                  & 135.8                  & Model             & \\Checkmark            & Scalar          \\
Shepherd           & 2853            & 3024.70                  & 404.64                  & Model             & \\Checkmark            & Scalar          \\
UltraFeedback         & 64015           & 651.31                   & 1161.04                 & 17 Models         & \\Checkmark            & Scalar/Textual  \\ \bottomrule
\end{tabular}
    
    }
    \caption{Statistics of existing critique datasets.}
    \label{tab:stat_preference_dataset}
\end{table*}
\fi

\subsection{\textsc{UltraFeedback}-Powered Models}
\label{sec:models}
Based on \textsc{UltraFeedback}, we develop UltraRM, an advanced open-source reward model that provides preferences for AI responses given user instructions. Additionally, we train a critique model UltraCM from the textual feedback in \textsc{UltraFeedback}. UltraCM could interact with human and AI assistants more flexibly in text. 

\looseness=-1
\textbf{UltraRM.} 
For reward modeling, we train UltraRM based on LLaMA2-13B~\citep{touvron2023llama}. Specifically, we train three versions of UltraRM.
We mix several open-source datasets with \textsc{UltraFeedback} to train UltraRM. The open-source datasets include Stanford SHP \cite{pmlr-v162-ethayarajh22a}, OpenAI Summarization~\citep{stiennon2020learning}, and Anthropic Helpful~\citep{bai2022training}. 
%In training the first two versions, we average the preference scores in each aspect to get a final reward score.  
% Recapping the annotation process, we have two granularity of scalar reward: (1) fine-grained scores that are annotated in preference annotation and inclined to evaluate model outputs from four aspects respectively, and (2) overall assessment ratings of model completion obtained in the critique generation process.
To validate the quality of UltraFeedback, we also train one model with \textit{merely} the fine-grained scores of this dataset, i.e. averaging the preference scores in each aspect to get a final reward score.
Further, to compare the effectiveness of the fine-grained scores and overall scores, we replace the fine-grained scores in UltraRM with the assessment ratings in critique generation, while remaining the open-source datasets.
% one reward model based on each kind of reward respectively and obtain two subsequent reward models: UltraRM and UltraRM-Overall.
The details for dataset processing can be found in Appendix~\ref{sec:app_data}.
We keep the training strategy, including loss objective and training hyperparameters, exactly the same as \citet{touvron2023llama}.

%Specifically, we set the margins to 0 for pairs from OpenAI Summarize and Anthropic Helpful, since there is no score annotation but merely binary preference labels. Given that the margin in UltraFeeback is within $\left(1, 5\right]$ whereas the range of scores in Stanford SHP is much wider, we normalize the margins to $\left(0, 1 \right]$ to avoid training instability due to a mismatch between the score scales of the two datasets.

\looseness=-1
\textbf{UltraCM.} 
We also train a critique model stemming from \textsc{UltraFeedback} to boost future research in learning from textual feedback \citep{wang2023mint}. 
UltraCM has the same initialization as UltraRM but is trained solely on \textsc{UltraFeedback}   critique data, i.e. $255,864$ textual feedback in total. 
Given a response, we fine-tune the model to give a corresponding critique that judges the response, figures out flaws, and provides suggestions for improvement. 
% We adopt the vanilla next token prediction task to perform supervised fine-tuning.

\vspace{-10pt}
\section{Experiments}
\label{sec:exp}
To validate the effect of AI feedback, we first evaluate UltraRM on human preference benchmarks in Section~\ref{sec:rm}. Next, we test UltraRM in enhancing chat language models with two strategies, namely best-of-$n$ sampling (Section~\ref{sec:bon}) and reinforcement learning (Section~\ref{sec:ppo}). 
Finally, we evaluate the feedback quality of UltraCM in Appendix~\ref{sec:cm}.
\vspace{-10pt}
\subsection{Reward Modeling}

\label{sec:rm}
\begin{table*}[thb]
\centering
\caption{Reward modeling accuracy (\%) results. We compare our UltraRM with baseline open-source reward models. LLaMA2 results are taken from \cite{touvron2023llama}. The highest results are in \textbf{bold} and the second highest scores are \underline{underlined}.}
\resizebox{0.9\textwidth}{!}{%
\begin{tabular}{@{}l|cc|ccccc@{}}
\toprule
\textbf{Model} & \textbf{Backbone Model} & \textbf{Open?} &\textbf{\begin{tabular}[c]{@{}c@{}}Anthropic\\      Helpful\end{tabular}} & \textbf{\begin{tabular}[c]{@{}c@{}}OpenAI\\      WebGPT\end{tabular}} & \textbf{\begin{tabular}[c]{@{}c@{}}OpenAI \\      Summ.\end{tabular}} & \textbf{\begin{tabular}[c]{@{}c@{}}Stanford\\      SHP\end{tabular}} & \textbf{Avg.} \\ \midrule
\textbf{Moss} & LLaMA-7B & $\checkmark$ & 61.3 & 58.1 & 59.0 & 54.6 & 58.3 \\
\textbf{Ziya} & LLaMA-7B & $\checkmark$ & 61.4 & 61.8 & 60.3 & 57.0 & 60.1 \\
\textbf{OASST} & DeBERTa-v3-large & $\checkmark$ & 67.6 & - & 71.8 & 53.9 & - \\
\textbf{SteamSHP} & FLAN-T5-XL & $\checkmark$ & 55.4 & 62.6 & 48.4 & 51.6 & 54.5 \\
\textbf{LLaMA2 Helpfulness} & LLaMA2-70B & \ding{55} & \textbf{72.0} & - & \textbf{75.5} & \textbf{80.0} & - \\
\midrule
\textbf{UltraRM} & LLaMA2-13B & $\checkmark$ & \underline{71.0} & \textbf{65.2} & \underline{74.0} & \underline{73.7} & \textbf{71.0} \\ 
\textbf{w/ Only \textsc{UltraFeedback}} & LLaMA2-13B & $\checkmark$ & 66.7 & 65.1 & 66.8 & 68.4 & 66.8 \\
\textbf{w/ Overall Score} & LLaMA2-13B & $\checkmark$ & \underline{71.0} & 62.0 & 73.0 & 73.6 & \underline{69.9} \\ \bottomrule
\end{tabular}
}
\vspace{-10pt}
\label{tab:reward}
\end{table*}

\textbf{Setup.} To evaluate how UltraRM aligns with human preference, we conduct experiments on four \textbf{human annotated} preference datasets, OpenAI WebGPT~\citep{nakano2021webgpt}, OpenAI Summarization~\citep{stiennon2020learning}, Anthropic HH-RLHF~\citep{bai2022training}, and Standford SHP. %Specifically, we train two versions of reward models, where UltraRM-UF is merely fine-tuned on \textsc{UltraFeedback} and UltraRM is fine-tuned on a mixture of \textsc{UltraFeedback} and an equal-size sample from three open-source datasets (except OpenAI WebGPT). 
On each dataset, we calculate the rewards of two responses for one prompt and predict which one is more preferred. We compare our UltraRM-UF, UltraRM-Overall, and UltraRM with open-source baselines, including Moss~\citep{zheng2023secrets}, Ziya~\citep{Fengshenbang-LM}, OASST~\footnote{\url{https://huggingface.co/OpenAssistant/reward-model-deberta-v3-large-v2}}, 
and SteamSHP \cite{pmlr-v162-ethayarajh22a}. We also report the results in LLaMA2~\citep{touvron2023llama}, although their reward models are not released. 

\looseness=-1
\textbf{Results.}  The preference prediction accuracy results are reported in Table~\ref{tab:reward}. As we can see, the UltraRM series outperforms baseline reward models except for the closed LLaMA2 reward model (much larger) by a large margin, indicating that UltraRM series are the \textbf{best open-source reward models}. Notably, our reward model can still surpass all other baselines even without mixing open-source datasets. These results reveal that, \textsc{UltraFeedback} is highly consistent with human preference, and its high quality as well as diversity enable strong out-of-distribution generalization. On average, the model trained with only \textsc{UltraFeedback} outperforms open-source baseline models by over $6.3$ percent in accuracy, while mixing open-source datasets with overall scores and fine-grained scores of \textsc{UltraFeedback} achieves $3.1$ and $4.2$ percent more improvement respectively.

We highlight that the OpenAI WebGPT dataset has no training and test splits, and neither most baselines nor we train reward models on this dataset\footnote{The OASST and LLaMA2 Helpfulness reward model used WebGPT dataset for training. To prevent data leakage, we do not report their performance on WebGPT.}, making it a fair benchmark to evaluate the generalization ability of reward models. Obviously, UltraRM series are significantly better, reaching $2.6\%$ absolute points improvement over baselines. Another intriguing finding is that adding open-source datasets has a minor effect on the WebGPT dataset, which again proves the transferability advantage of \textsc{UltraFeedback}.
On another benchmark Stanford SHP, UltraRM also achieves remarkable performance.

A noteworthy finding is that, despite exhibiting comparably on the other three datasets, the reward model trained with overall scores discernably lags behind the other two variants on WebGPT. There can be two potential explanations for this observation. 
First, fine-grained annotation, which scores model outputs from different aspects respectively, provides a more precise assessment for each completion than aggregating evaluation into an overall number.
Second, in the overall quality annotation process, each sample is sent to GPT-4 separately whereas, in fine-grained rating, all four completions are scored at the same time, which may provide GPT-4 with cross-references and prevent it from applying inconsistent standards, reducing the impact of randomness. These superiorities demonstrate the high quality of our fine-grained preference data, and we advocate future work to adopt the fine-grained annotation schema and rate multiple completions at one time.

% \usepackage{booktabs}

\begin{table*}[thb]
\centering
\caption{Head-to-head comparison results on three public benchmarks. The baseline is \texttt{text-davinci-003} in AlpacaEval and \texttt{gpt-3.5-turbo} in Evol-Instruct and UltraChat. The judge is GPT-4. The highest win rates are in \textbf{bold}.}
\resizebox{0.9\textwidth}{!}{%
\begin{tabular}{@{}lccccc@{}}
\toprule
\textbf{Model} & \textbf{Size} & \begin{tabular}[c]{@{}c@{}}\textbf{AlpacaEval} \\ \textbf{Win (\%)}\end{tabular} & \begin{tabular}[c]{@{}c@{}}\textbf{Evol-Instruct} \\ \textbf{Win / Tie / Lose (\%)}\end{tabular} & \begin{tabular}[c]{@{}c@{}}\textbf{UltraChat} \\ \textbf{Win / Tie / Lose (\%)}\end{tabular} & \begin{tabular}[c]{@{}c@{}}\textbf{Average} \\ \textbf{Win (\%)}\end{tabular} \\
\midrule
\textbf{ChatGPT} & - & 89.4 & - & - & - \\ \midrule
\multicolumn{6}{c}{\textit{LLaMA2}} \\ \midrule
\textbf{Vicuna-13B-v1.5} & 13B & - & 33.0 / 23.9 / 43.1 & 34.5 / 38.2 / 27.3 & - \\
\textbf{LLaMA2-13B-Chat} & 13B & 81.1 & 44.1 / 11.9 / 44.0 & 53.5 / 21.3 / 25.2 & 59.5 \\
\textbf{WizardLM-13B-v1.2} & 13B & 89.2 & 55.5 / 17.4 / 27.1 & 59.7 / 25.5 / 14.8 & 68.1 \\
\textbf{OpenChat-13B-v3.2super} & 13B & 89.5 & 55.5 / 11.0 / 33.5 & 58.7 / 26.7 / 14.5 & 67.9 \\
\textbf{LLaMA2-70B-Chat} & 70B & \textbf{92.7} & 56.4 / 13.8 / 29.8 & 54.0 / 28.6 / 17.4 & 67.7 \\ \midrule
\multicolumn{6}{c}{\textit{LLaMA}} \\ \midrule
\textbf{UltraLM-13B} & 13B & 80.7 & 39.9 / 14.7 / 45.4 & 38.2 / 34.8 / 27.0 & 52.9 \\
\textbf{Vicuna-13B-v1.3} & 13B & 82.1 & 36.7 / 17.4 / 45.9 & 41.3 / 33.2 / 25.5 & 53.4 \\
\textbf{WizardLM-13B-v1.1} & 13B & 86.3 & 54.1 / 14.7 / 31.2 & 56.1 / 26.0 / 17.9 & 65.5 \\
\textbf{Vicuna-33B-v1.3} & 33B & 89.0 & 50.0 / 17.0 / 33.0 & 57.7 / 25.7 / 16.6 & 65.6 \\
\textbf{UltraLM-13B-PPO} & 13B & 86.3 & \textbf{57.8} / 10.1 / 32.1  & \textbf{64.9} / 15.6 / 19.5 & \textbf{69.7}\\
\bottomrule
\end{tabular}
}
\vspace{-10pt}
\label{tab:exp_res}
\end{table*}

\subsection{Best-of-$n$ Experiments}
\label{sec:bon}
% \begin{wrapfigure}{R}{0.5\textwidth}
% \vspace{-15pt}
%   \begin{center}
%     \includegraphics[width=0.5\textwidth]{figures/bon.pdf}
%   \end{center}
%   \vspace{-15pt}
%   \caption{Win rate against \texttt{text-davinci-003} on AlpacaEval benchmark. We sample $n$ responses and choose the one with the highest reward.}
%   \label{fig:bon}
% \vspace{-20pt}
% \end{wrapfigure}
\begin{figure}[thb]
% \vspace{-15pt}
  \begin{center}
    \includegraphics[width=.9\linewidth]{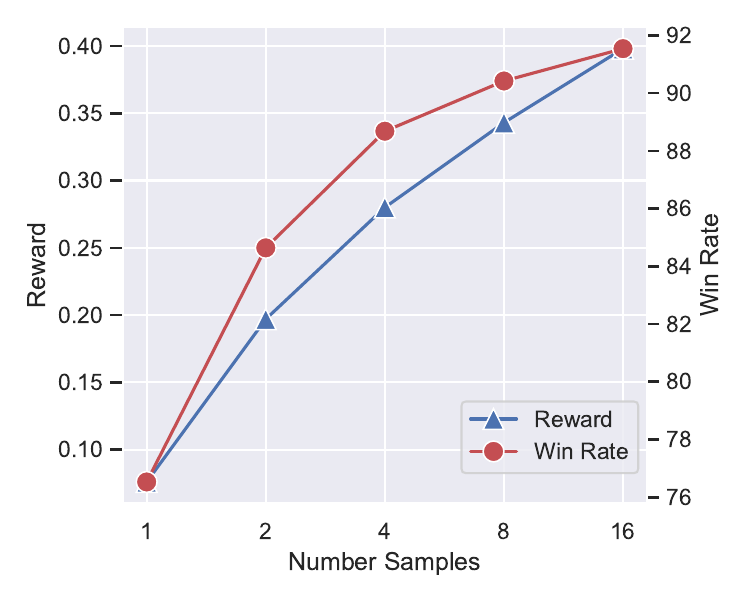}
  \end{center}
  \vspace{-15pt}
  \caption{Win rate against \texttt{text-davinci-003} on AlpacaEval benchmark. We sample $n$ responses and choose the one with the highest reward.}
  \label{fig:bon}
\vspace{-10pt}
\end{figure}

\textbf{Setup.} To verify that UltraRM could serve as a good indicator of response quality, we conduct best-of-$n$ experiments. On the AlpacaEval benchmark, we randomly sample $16$ completions from the original UltraLM-13B and calculate their corresponding rewards. We then select the best-of-$\{1,2,4,8,16\}$ completions as the final response. The sampling parameters are set to $\text{temperature}=1$ and top-$p=1$.

\textbf{Results.} We present results in Figure~\ref{fig:bon}. Apparently, we can see the win rate on AlpacaEval increases proportionally with rewards. This validates that our UltraRM gives rigorous rewards that reflect the overall response quality. Notably, the best-of-$n$ sampling strategy is surprisingly effective. The initial UltraLM-13B model achieves a $76.53\%$ win rate for a single sampling, and a simple best-of-$2$ sample increases the win rate to $84.64\%$. With more samples, we can get even more high-quality responses, and the final best-of-$16$ win rate hits $91.54\%$. The best-of-$n$ sampling is universally applicable across models and tasks, which enhances models without training. Please refer to Appendix~\ref{sec:case_bon} for cases.

\subsection{PPO Experiments}
\label{sec:ppo}

\textbf{Setup.} Given the state-of-the-art UltraRM, we aim to push the upper bound of open-source chat language models with RLAIF. Specifically, we perform PPO over UltraLM-13B~\citep{ding2023enhancing} to get its PPO version, UltraLM-13B-PPO. We tune UltraLM for 80 iterations on the \textsc{UltraFeedback} prompts. In each iteration, we collect 512 samples and update the policy model with a mini-batch size of 64. The learning rate is fixed at 1e-6.

\textbf{Baselines.} We compare UltraLM-13B-PPO with leading open-source models and proprietary models, including LLaMA2-Chat~\citep{touvron2023llama}, Vicuna~\citep{vicuna2023}, WizardLM~\citep{xu2023wizardlm},
OpenChat~\citep{wang2023openchat}, and ChatGPT~\citep{openai2022chatgpt}.

\textbf{Benchmarks.} We conduct experiments on three public benchmarks, namely AlpacaEval~\citep{alpaca_eval}, Evol-Instruct~\citep{xu2023wizardlm}, and UltraChat~\citep{ding2023enhancing}. On each benchmark, we ask GPT-4 to judge which response is better given the same instruction.
AlpacaEval adopts \texttt{text-davinci-003} as the competitor model, while we compete with \texttt{gpt-3.5-turbo} on Evol-Instruct and UltraChat.
To avoid position bias, we randomly switch the comparing responses. For all models, we use the same decoding parameter with $\text{temperature}=0.7$ and top-$p=1$.

\textbf{Results.} We report experiment results in Table~\ref{tab:exp_res}. We take the official results on the AlpacaEval leaderboard for baseline models and conduct evaluations by ourselves for other results. Overall, our UltraLM-13B-PPO achieves the highest average win rate on the three benchmarks, outperforming all other open-source models. Among LLaMA-based models, UltraLM-13B-PPO overtakes other models by at least $3.6$ percent on average. Even when compared with the much larger LLaMA2-70B-Chat model, our model still holds the advantage, illustrating the huge benefit of RLAIF alignment.
Our model also reaches the highest win rate on two of the benchmarks, Evol-Instruct and UltraChat, against the more powerful \texttt{gpt-3.5-turbo}.
It is worth noting that, compared with the original UltraLM-13B, the PPO process benefits the model greatly, leading to a $16.8$ percent enhancement. We provide cases in Appendix~\ref{sec:case_ppo}.%On the Evol-Instruct and UltraChat test set, we also evaluate the relative scores compared with \texttt{gpt-3.5-turbo}. As presented in Table~\ref{tab:evol}, UltraLM-13B-PPO still gets the highest score, outperforming all strong baselines. 

\vspace{-10pt}
\section{Agreement with Human Preferences}
\label{sec:human}
The inclusivity of human preferences is known to be hard to capture~\citep{dubois2023alpacafarm}. Heavily relying on AI feedback, it is essential to measure and monitor the agreement between AI and human preferences. In this section, we conduct experiments to see (1) to what extent AI annotations are consistent with human preferences (Section~\ref{sec:human_con}) and (2) how reliable AI evaluations are (Section~\ref{sec:human_eval}).
%The annotation and evaluation of \textsc{UltraFeedback} heavily relies on GPT-4, which is undoubtedly the most advanced AI system so far. 
%Although many researchers pointed out that GPT-4 is capable of providing meticulous judgments~\citep{dubois2023alpacafarm,zheng2023judging} and feedback~\citep{ bai2022constitutional, lee2023rlaif}, and exhibits high agreement with human preference, human preference itself could still be hard to 
\iffalse
\begin{table}[]
\caption{Agreement between different judges on 400 samples from \textsc{UltraFeedback}, AlpacaEval, Evol-Instruct and UltraChat test sets (100 each). A-1, A-2, A-3 are three human judges. We include tie votes and the random agreement is 33\%.}
\begin{tabular}{@{}lccccc@{}}
\toprule
\textbf{Judge} & \textbf{GPT-4} & \textbf{A-1} & \textbf{A-2} & \textbf{A-3} & \textbf{Average} \\ \midrule
\textbf{GPT-4} & -              & 59.2\%       & 60.8\%       & 59.1\%       & \textbf{59.7\%}           \\
\textbf{A-1}   & 59.2\%         & -            & 58.1\%       & 54.7\%       & 57.3\%           \\
\textbf{A-2}   & 60.8\%         & 58.1\%       & -            & 55.4\%       & 58.1\%           \\
\textbf{A-3}   & 59.1\%         & 54.7\%       & 55.4\%       & -            & 56.4\%           \\ \bottomrule
\end{tabular}
\label{tab:human_agreement}
\end{table}
\fi

\subsection{Annotation Consistency}
\label{sec:human_con}
\looseness=-1
In Section~\ref{sec:rm}, we show that the reward models trained on \textsc{UltraFeedback} could predict human preference accurately.
To further analyze to what extent AI feedback could capture human preference, we randomly sample $400$ comparison pair from \textsc{UltraFeedback}, AlpacaEval, Evol-Instruct and UltraChat test sets (100 each) and ask 3 independent annotators to compare those pairs (win/tie/lose). The annotators are undergraduate and graduate students. We present the agreement ratio between GPT-4 and annotators, as well as annotators themselves in Table~\ref{tab:human_agreement}. On average, GPT-4 judge exhibits 59.7\% agreement rate with human labelers, which matches previous human evaluation on MT-Bench~\citep{zheng2023judging}. We also observe similar agreement rates among annotators. Notably, the agreement between GPT-4 and the majority votes of three annotators raises to 68.6\%, meaning that GPT-4 better reflects the collective human preferences.
%We conduct a probing experiment with $100$ random comparison pairs from \textsc{UltraFeedback}. The agreement between human and GPT-4 annotations on the $100$ samples is $74.2\%$. 
%According to previous research~\citep{ouyang2022training, dubois2023alpacafarm}, the agreement rate between qualified human annotators is also around $70\%$. Therefore, the probing experiment (roughly) validates that \textsc{UltraFeedback} is aligned with humans.
\begin{table}[thb]
\vspace{-13pt}
\caption{Agreement between judges on 400 samples from \textsc{UltraFeedback}, AlpacaEval, Evol-Instruct, and UltraChat test sets . A-1, A-2, A-3 are three human judges. ``Majority'' stands for the agreement between each judge and other three judges's majority votes. We include tie votes and the random agreement is 33\%.}
% \vspace{5pt}
\resizebox{\linewidth}{!}{
\begin{tabular}{@{}lccccc@{}}
\toprule
\textbf{Judge} &  \textbf{A-1} & \textbf{A-2} & \textbf{A-3} & \textbf{Average} & \textbf{Majority}\\ \midrule
\textbf{GPT-4} &  59.2\%       & 60.8\%       & 59.1\%       & \textbf{59.7\%}      & \textbf{68.6\%}     \\
\textbf{A-1}          & -            & 58.1\%       & 54.7\%       & 57.3\%      & 60.3\%     \\
\textbf{A-2}    & 58.1\%       & -            & 55.4\%       & 58.1\%      & 63.3\%     \\
\textbf{A-3}      & 54.7\%       & 55.4\%       & -            & 56.4\%      & 62.0\%     \\ \bottomrule
\end{tabular}
}
\vspace{-17pt}
\label{tab:human_agreement}
\end{table}

\subsection{Reliability of AI Evaluation}
\label{sec:human_eval}
We first supplement another AI evaluation using Claude-3 Sonnet \cite{claude3} to investigate the agreement among different series of AI models. The prompts are the same as the GPT-4 evaluation.
Then, we compare both of our AI evaluation results with human annotations to examine if AI evaluations reliably correlate with humans. Particularly, we use the majority votes of the three annotators and filter out samples with all different votes. 

We present GPT-4, Claude-3, and human evaluation results on the remaining 266 pairs in Table~\ref{tab:app_human}. Overall, Claude-3 shares the same trend as GPT-4 and further increases our models' win rates.
Human evaluations are mostly consistent with GPT-4, giving a 64.3\% against 67.3\% average winning rate. We notice that human labelers tend to assign more ties than GPT-4, leading to slightly lower winning rates.

\begin{figure}[thb]
    % \vspace{-10pt}
    \centering
    \includegraphics[width=\linewidth]{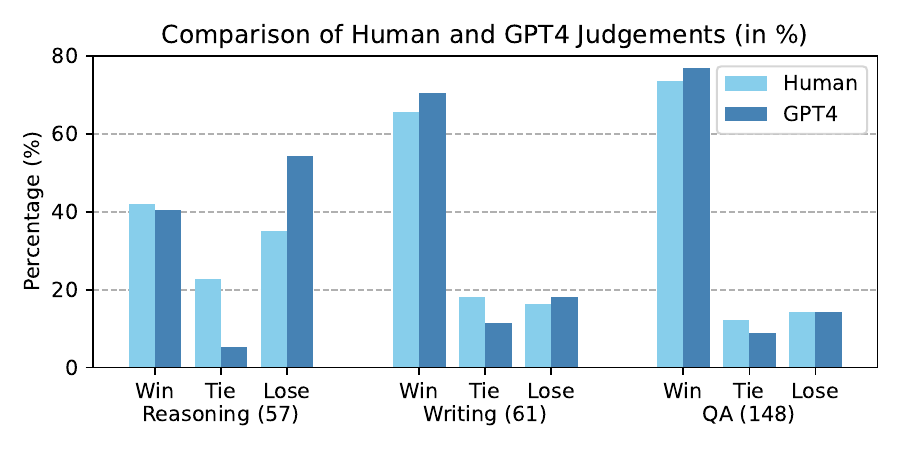}
    \vspace{-20pt}
    \caption{Catrgorical comparison of human and GPT-4 judgments. Human judgments are majority votes from three annotators. Sample numbers of each category are in parentness.}
    \vspace{-10pt}
    \label{fig:human_cate}
\end{figure}

For fine-grained analysis, we categorize the evaluated samples into reasoning, writing, and QA tasks. The categorical comparison results are presented in Figure~\ref{fig:human_cate}. It is shown that human evaluations are mostly consistent with GPT-4, where they both prefer our models on writing and QA tasks. On reasoning tasks including coding, math, and logic, human and GPT-4 judgments diverge on ties and losses, where GPT-4 gives fewer ties but more losses. To delve into the discrepancy deeper, we ask another expert labeler to determine the ground truth answer for each question. In this sense, a model wins when it gives the correct answer while the other does not, and vice versa. The two models tie when they both successfully or unsuccessfully answer the question. The final win/tie/lose rate comes at $42.1 \%/ 26.3\% / 31.6\%$ and closely matches human evaluations. 
The GPT-4 judge, in this case, potentially underestimated our model's reasoning performance and still has space to improve.
%Against \texttt{text-davinci-003}, human annotators gave our model even higher win rate. Against \texttt{gpt-3.5-turbo}, human annotators tended to assign more ties and less wins, and the lose rate of our model slightly increases.

%We have conducted human evaluation on UltraLM-13B-PPO. Specifically, we hire 5 independent annotators to compare model responses. We randomly sample 100 comparison pairs from each benchmark (300 in total) and ask annotators to rank the responses (win/tie/lose). We present human and GPT-4 evaluation results on the 300 pairs in Table~\ref{tab:app_human}. Overall, human evaluations are mostly consistent with GPT-4 annotations. Against \texttt{text-davinci-003}, human annotators gave our model even higher win rate. Against \texttt{gpt-3.5-turbo}, human annotators tended to assign more ties and less wins, and the lose rate of our model slightly increases.
\vspace{-15pt}
\begin{table}[thb]
\centering
\caption{Human evaluation results. We use majority votes from three human judges and compare GPT-4, Claude-3, and human evaluations on the same 266 samples.}
\resizebox{\linewidth}{!}{
\begin{tabular}{@{}lcccc@{}}
\toprule
\multirow{2}{*}{\textbf{Judge}} & 
\textbf{AlpacaEval}  & \textbf{Evol-Instruct} & \textbf{UltraChat} & \textbf{Avg.} \\
& \textbf{Win (\%)} & \multicolumn{2}{c}{\textbf{Win / Tie / Lose (\%)}} & \textbf{Win (\%)} \\
\midrule
\textbf{GPT-4}                    & 83.9                      & 57.1 / 8.8 / 34.1    & 61.0 / 17.1 / 21.9   & 67.3  \\
\textbf{Claude-3}                 & 95.1
    & 59.6 / 1.4 / 39.0     & 73.5 / 6.8 / 19.7 & 76.1 \\
\textbf{Human}                    & 78.5                      & 68.1 / 17.6 / 14.3    &   46.3 / 19.5 / 34.1  & 64.3 \\ \bottomrule
\end{tabular}
}
\vspace{-10pt}
\label{tab:app_human}
\end{table}
\vspace{-5pt}
\vspace{-5pt}
\section{Analysis}
\label{sec:analysis}

In this section, we further analyze how \textsc{UltraFeedback} enhances language models on different subjects (Section~\ref{sec:type}) and tasks (Section~\ref{sec:capa}).
%To further understand how \textsc{UltraFeedback} enhances language models, we conduct a finer analysis. In Section~\ref{sec:type}, we observe the type-specific performances.

\subsection{Question Type Breakdown}
\label{sec:type}
\begin{figure*}[thb]
    \centering
    \includegraphics[width=\linewidth]{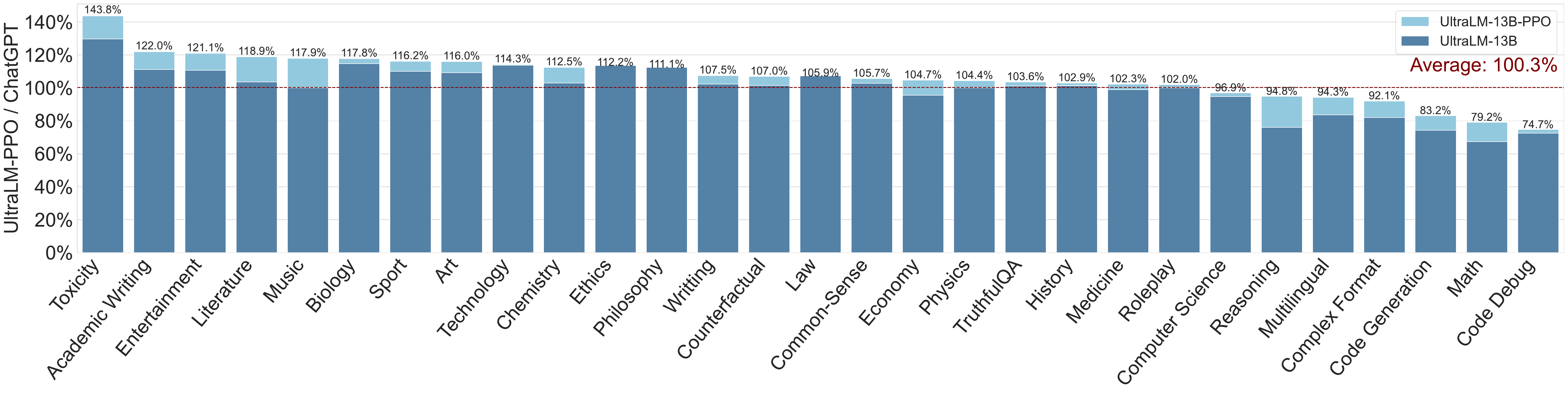}
    \vspace{-25pt}
    \caption{Comparison results between UltraLM-13B-PPO, UltraLM-13B, and \texttt{gpt-3.5-turbo} on Evol-Instruct test set, where \texttt{gpt-3.5-turbo} scores are $100\%$.}
    \vspace{-15pt}
    \label{fig:evol}
\end{figure*}

\begin{table*}[thb]
\caption{Exact match scores (\%) for UltraLM-13B and UltraLM-13B-PPO on model capability benchmarks.}
\resizebox{\textwidth}{!}{%
\begin{tabular}{@{}l|ccccccccccc|c@{}}
\toprule
\textbf{Model} & \textbf{BoolQ} & \textbf{HellaSwag} & \textbf{RACE-h} & \textbf{RACE-m} & \textbf{MultiRC} & \textbf{TriviaQA} & \textbf{NQ} & \textbf{PIQA} & \textbf{OBQA} & \textbf{ARC-E} & \textbf{ARC-C} & \textbf{Avg.} \\ \midrule
\textbf{UltraLM-13B} & 85.0 & 59.8 & 66.1 & 73.5 & 83.2 & 50.8 & 19.4 & 73.5 & 57.0 & 76.1 & 51.5 & 63.3 \\
\textbf{UltraLM-13B-PPO} & 83.5 & 62.6 & 66.8 & 74.2 & 83.7 & 52.5 & 22.1 & 74.9 & 57.0 & 76.1 & 53.9 & 64.3 \\
\bottomrule
\end{tabular}}
\vspace{-15pt}
\label{tab:static}
\end{table*}

Figure~\ref{fig:evol} reports the UltraLM-13B-PPO and UltraLM-13B scores on different question types versus \texttt{gpt-3.5-turbo} on the Evol-Instruct test set. We observe that UltraLM-13B-PPO overtakes ChatGPT on 22/29 subjects, especially on writing-related tasks such as academic writing. Our model is also well-aligned with human values, getting higher scores on toxicity, ethics, and TruthfulQA. On some difficult subjects like roleplay, reasoning, and counterfactual, our model is still on par with ChatGPT, indicating the strong advanced model capability.
Compared with the original UltraLM-13B, PPO boosts the model in multiple aspects, including professional knowledge (economy, chemistry, music, literature) and reasoning ability (reasoning, complex format, code generation, math). 
Meanwhile, our model falls behind \texttt{gpt-3.5-turbo} on math and code-related tasks, which might be attributed to the limitation of base model ability and the lack of relevant data in \textsc{UltraFeedback}. Table~\ref{tab:ultra} in Appendix~\ref{sec:app_type} provides additional results on the UltraChat test set and reaches the same conclusion. We leave this as our future work.

%which are important directions in our future work.%Specifically, the RLHF version of UltraLM-13B improves by $6.6$ percent over the original UltraLM-13B. Also, our model surpasses the powerful ChatGPT and WizardLM series, which was trained on the Evol-Instruct training set. 

\subsection{Does RLAIF Benefit Model Capability?}
\label{sec:capa}
To test whether RLAIF impacts base model capability, we conduct experiments on nine commonly used benchmarks including question answering and multiple-choice questions (See Appendix~\ref{sec:app_capa} for details). We compare UltraLM-13B before and after PPO. The results in Table~\ref{tab:static} demonstrate marginal improvements over these benchmarks with about 1 absolute point. We note that this is in line with established conclusions \citep{openai2023gpt4} regarding RLHF, which state that RLHF could produce more preferable responses, but has a minor effect on model capability.

\vspace{-10pt}

\section{Related Work}
\paragraph{Feedback Learning for LLMs.}
Incorporating human feedback with imitation learning or reinforcement learning~\citep{schulman2017proximal, rafailov2023direct} has been the mainstream approach to align LLMs with human preferences in leading cooperations~\citep{stiennon2020learning, ouyang2022training, bai2022training, glaese2022improving, openai2022chatgpt, openai2023gpt4, touvron2023llama1}. 
However, human feedback relies on human capabilities, which makes it hard to scale up and apply to superhuman tasks. Accordingly, some researchers proposed \textit{scalable oversight}, which aims to supervise potent AI models by models themselves~\citep{Irving2018AISV, Leike2018ScalableAA, Christiano2018SupervisingSL}. Empirically for LLMs, \citet{bai2022constitutional} first presented Constitutional AI to let LLMs refine their responses given a set of regulations. \citet{lee2023rlaif} and \citet{Burns2023WeaktoStrongGE} further validated that learning from AI feedback could surpass human feedback on some specific tasks. More broadly, our work verified that scaled AI feedback could enhance the general ability of open-source chat models.
%Another line of research focuses on 
%In the past few years, reinforcement learning for language models has been extensively studied by leading corporations such as OpenAI, Anthropic, and Google DeepMind. Early works applied RLHF to enhance language models on specific tasks such as summarization~\citep{stiennon2020learning} and question answering~\citep{nakano2021webgpt, menick2022teaching}. Afterward, RLHF has been widely adopted on advanced general language models~\citep{ouyang2022training, bai2022training, glaese2022improving, openai2022chatgpt, openai2023gpt4, touvron2023llama} to align with human values, and the RLHF versions are mostly more helpful, more truthful, and less harmful~\citep{perez2022red,ouyang2022training}. However, these top companies seldom open-source their preference datasets and reward models, hindering other researchers from reproducing their results. The open-source research community also put efforts into demystifying RLHF, which dive into the RL implementation details~\citep{zheng2023secrets, song2023reward}, design novel algorithms ~\citep{gulcehre2023reinforced, rafailov2023direct, song2023preference} to utilize preference data more efficiently, or align LLMs with various types of feedbacks~\citep{yao2023retroformer, wang2023shepherd, liu2023rltf}. To compare, our work explores \textbf{how to automatically construct high-quality feedback data}, which is fundamental yet usually overlooked in the literature.
\vspace{-10pt}
\paragraph{Data for LLM Alignment.}
The importance of data scalability and quality has been widely recognized in the literature on instruction tuning (also known as SFT). Early works collected various NLP tasks or real user conversations to conduct instruction tuning and observed that LLMs could generalize well across different tasks~\citep{wei2022finetuned,sanh2022multitask,wang2022super,ouyang2022training}. After the release of ChatGPT, most recent research on SFT emphasized the importance of data construction and reached conclusions that scalability, diversity, as well as quality, are vital for the final performance~\citep{ding2023enhancing, taori2023stanford, vicuna2023, xu2023wizardlm}. However, when it goes to the feedback learning stage, the importance of data engineering has not been well illustrated. Among current preference datasets, some of them focus on specific tasks (e.g. summarization~\citep{stiennon2020learning}, search-based question answering~\citep{nakano2021webgpt}, safety-oriented scenarios~\citep{ji2023beavertails}, and math problems~\citep{lightman2023let}), thus cannot boost general chat models. Some datasets are small in scale~\citep{wu2023fine, wang2023shepherd} or provide only community votes as coarse-grained preferences~\citep{pmlr-v162-ethayarajh22a, askell2021general}. Therefore, a large general-purpose preference dataset with diverse instructions and fine-grained annotations in the open-source community is urgently in need, which motivates us to construct \textsc{UltraFeedback}.

%OpenAI opened their preference datasets on summarization~\citep{stiennon2020learning}, search-based question answering~\citep{nakano2021webgpt}, and math problems~\citep{lightman2023let} but did not release more general datasets or reward models. 
%To make language models follow human instructions and align with human values, instruction tuning (also known as SFT) was proposed and broadly applied~\citep{wei2022finetuned,sanh2022multitask,wang2022super,ouyang2022training} on LLMs. In the open-source community, 
%Current preference datasets are either task-specific~\citep{stiennon2020learning, nakano2021webgpt}, Some datasets are safety-oriented and 

\vspace{-10pt}
\section{Conclusion}
In this paper, we proposed to enhance open-source LLMs with scaled AI feedback. Through meticulous designation, we constructed \textsc{UltraFeedback}, a large-scale and diverse AI feedback dataset. With the data, we embarked on a thorough exploration of AI feedback's multifaceted utilities, including modeling human preferences, improving chat language models, and training critique models.
Our analysis further delved deep into human agreement and model capability evaluations, revealing some nuanced insights.
We believe that AI feedback would become a scalable and reliable source for future AI oversight. We hope our work could serve as an early exploration and data support in this area, facilitating researchers in the open-source community. 
%further revealed the high consistency between human and AI preferences, with a category-level investigation of the effect of RLAIF. 
In future work, we will continue exploring diverse, high-quality, and scalable preference data construction, expanding AI feedback in multi-turn dialogues, complex reasoning, coding, and safety scenarios.
%, a large and diverse preference dataset with fine-grained annotations in various formats. \textsc{UltraFeedback} has multiple usages, including training reward models for RLHF and training critique models for automatic evaluation and model interactions. Our experiments demonstrated the excellence of \textsc{UltraFeedback} and its potential to assist future feedback-learning research. We hope \textsc{UltraFeedback} could become a strong support and data basis in this area, facilitating researchers in the open-source community. 
% \newpage
\section*{Impact Statement}
Aligning AI systems, especially advanced LLMs, is important for the safety and trustworthiness in their applications.
We manage to enhance open LLMs with scaled AI feedback which is an underexplored research direction. With high efficiency and low cost, leveraging AI feedback could significantly reduce the consumation of human labors, leading to more scalable alignment.
We should also raise attention to the limitations of AI feedback, LLMs could be biased towards certain features, such as answer positions~\citep{zheng2023judging}, response lengths, and certain styles. In this way, such biases might lead to inaccurate or unfair annotations and evaluations. By overcoming these biases, more precise and helpful AI feedback can be obtained.  
In terms of \textsc{UltraFeedback}, we could expect it to improve a considerable amount of open-source LLMs and narrow their gaps with close-sourced models. 
We did not add safety-oriented conversations intentionally, so there could still be toxicity and unethical behaviors in the aligned models if prompted adversarially. We believe our paradigm is still useful for enhancing model safety, and are extensively working on it.
%Also, the annotations in \textsc{UltraFeedback} heavily rely on GPT-4. As all existing LLMs do, GPT-4 may also generate hallucinations and falsehoods. This issue also exists for our UltraLM-13B-PPO and UltraCM models. In the 
%We will eagerly dive into  it only contains general-purpose conversational instructions and  As a result, there could still be toxicity and unethical behaviors in our released models if prompted adversarially. 
Alongside data, we also release a series of models for feedback learning research. The reward model and critique model can be directly used to align LLMs for more prefered behaviors. On the other hand, although our models are potent in solving tasks and giving feedback, they may also generate hallucinations and falsehoods. The risk of misuse is a severe threat to open LLMs, which calls for appropriate regulation and supervision.

\iffalse
\section*{Reproducibility Statement}
We upload \textsc{UltraFeedback} anomalously to this \href{https://www.dropbox.com/scl/fi/5v3gl4igxjdqi5qdhge54/UltraFeedback.zip?rlkey=6y2vu2yrfdzer14b6gumjnsqg&dl=0}{link}. 
% \textsc{UltraFeedback} dataset is released at this \href{https://huggingface.co/datasets/openbmb/UltraFeedback}{link}. 
To reproduce reward modeling and critique modeling experiment results, please download it and follow the training parameters and data processing details in Section~\ref{sec:models} to train UltraRM and UltraCM, and evaluate them accordingly on the benchmark datasets.
% To reproduce the best-of-$n$ and reinforcement learning results, please get UltraLM from this \href{https://huggingface.co/openbmb/UltraLM-13b}{link}, then follow the instructions in Section~\ref{sec:bon} and~\ref{sec:ppo}.
\fi

\section*{Acknowledgement}
This work is supported by the National Key R\&D Program of China (No.2022ZD0116312), National Natural Science Foundation of China (No. 62236004).

\bibliography{icml2024}

\begin{thebibliography}{73}
\providecommand{\natexlab}[1]{#1}
\providecommand{\url}[1]{\texttt{#1}}
\expandafter\ifx\csname urlstyle\endcsname\relax
  \providecommand{\doi}[1]{doi: #1}\else
  \providecommand{\doi}{doi: \begingroup \urlstyle{rm}\Url}\fi

\bibitem[Almazrouei et~al.(2023)Almazrouei, Alobeidli, Alshamsi, Cappelli,
  Cojocaru, Debbah, Goffinet, Heslow, Launay, Malartic, Noune, Pannier, and
  Penedo]{falcon40b}
Ebtesam Almazrouei, Hamza Alobeidli, Abdulaziz Alshamsi, Alessandro Cappelli,
  Ruxandra Cojocaru, Merouane Debbah, Etienne Goffinet, Daniel Heslow, Julien
  Launay, Quentin Malartic, Badreddine Noune, Baptiste Pannier, and Guilherme
  Penedo.
\newblock {Falcon-40B}: an open large language model with state-of-the-art
  performance.
\newblock 2023.

\bibitem[Anthropic(2024)]{claude3}
Anthropic.
\newblock Introducing the next generation of claude.
\newblock 2024.

\bibitem[Askell et~al.(2021)Askell, Bai, Chen, Drain, Ganguli, Henighan, Jones,
  Joseph, Mann, DasSarma, et~al.]{askell2021general}
Amanda Askell, Yuntao Bai, Anna Chen, Dawn Drain, Deep Ganguli, Tom Henighan,
  Andy Jones, Nicholas Joseph, Ben Mann, Nova DasSarma, et~al.
\newblock A general language assistant as a laboratory for alignment.
\newblock \emph{arXiv preprint arXiv:2112.00861}, 2021.

\bibitem[Austin et~al.(2021)Austin, Odena, Nye, Bosma, Michalewski, Dohan,
  Jiang, Cai, Terry, Le, et~al.]{austin2021program}
Jacob Austin, Augustus Odena, Maxwell Nye, Maarten Bosma, Henryk Michalewski,
  David Dohan, Ellen Jiang, Carrie Cai, Michael Terry, Quoc Le, et~al.
\newblock Program synthesis with large language models.
\newblock \emph{arXiv preprint arXiv:2108.07732}, 2021.

\bibitem[Bai et~al.(2022{\natexlab{a}})Bai, Jones, Ndousse, Askell, Chen,
  DasSarma, Drain, Fort, Ganguli, Henighan, et~al.]{bai2022training}
Yuntao Bai, Andy Jones, Kamal Ndousse, Amanda Askell, Anna Chen, Nova DasSarma,
  Dawn Drain, Stanislav Fort, Deep Ganguli, Tom Henighan, et~al.
\newblock Training a helpful and harmless assistant with reinforcement learning
  from human feedback.
\newblock \emph{arXiv preprint arXiv:2204.05862}, 2022{\natexlab{a}}.

\bibitem[Bai et~al.(2022{\natexlab{b}})Bai, Kadavath, Kundu, Askell, Kernion,
  Jones, Chen, Goldie, Mirhoseini, McKinnon, et~al.]{bai2022constitutional}
Yuntao Bai, Saurav Kadavath, Sandipan Kundu, Amanda Askell, Jackson Kernion,
  Andy Jones, Anna Chen, Anna Goldie, Azalia Mirhoseini, Cameron McKinnon,
  et~al.
\newblock Constitutional ai: Harmlessness from ai feedback.
\newblock \emph{arXiv preprint arXiv:2212.08073}, 2022{\natexlab{b}}.

\bibitem[Biderman et~al.(2023)Biderman, Schoelkopf, Anthony, Bradley, O'Brien,
  Hallahan, Khan, Purohit, Prashanth, Raff, Skowron, Sutawika, and van~der
  Wal]{Biderman2023pythia}
Stella Biderman, Hailey Schoelkopf, Quentin~Gregory Anthony, Herbie Bradley,
  Kyle O'Brien, Eric Hallahan, Mohammad~Aflah Khan, Shivanshu Purohit,
  USVSN~Sai Prashanth, Edward Raff, Aviya Skowron, Lintang Sutawika, and Oskar
  van~der Wal.
\newblock Pythia: {A} suite for analyzing large language models across training
  and scaling.
\newblock In Andreas Krause, Emma Brunskill, Kyunghyun Cho, Barbara Engelhardt,
  Sivan Sabato, and Jonathan Scarlett (eds.), \emph{International Conference on
  Machine Learning, {ICML} 2023, 23-29 July 2023, Honolulu, Hawaii, {USA}},
  Proceedings of Machine Learning Research, 2023.

\bibitem[Bisk et~al.(2020)Bisk, Zellers, Gao, Choi, et~al.]{bisk2020piqa}
Yonatan Bisk, Rowan Zellers, Jianfeng Gao, Yejin Choi, et~al.
\newblock Piqa: Reasoning about physical commonsense in natural language.
\newblock In \emph{Proceedings of the AAAI conference on artificial
  intelligence}, 2020.

\bibitem[Brown et~al.(2020)Brown, Mann, Ryder, Subbiah, Kaplan, Dhariwal,
  Neelakantan, Shyam, Sastry, Askell, et~al.]{brown2020language}
Tom Brown, Benjamin Mann, Nick Ryder, Melanie Subbiah, Jared~D Kaplan, Prafulla
  Dhariwal, Arvind Neelakantan, Pranav Shyam, Girish Sastry, Amanda Askell,
  et~al.
\newblock Language models are few-shot learners.
\newblock \emph{Advances in neural information processing systems},
  33:\penalty0 1877--1901, 2020.

\bibitem[Burns et~al.(2023)Burns, Izmailov, Kirchner, Baker, Gao,
  Aschenbrenner, Chen, Ecoffet, Joglekar, Leike, Sutskever, Wu, and
  OpenAI]{Burns2023WeaktoStrongGE}
Collin Burns, Pavel Izmailov, Jan~Hendrik Kirchner, Bowen Baker, Leo Gao,
  Leopold Aschenbrenner, Yining Chen, Adrien Ecoffet, Manas Joglekar, Jan
  Leike, Ilya Sutskever, Jeff Wu, and OpenAI.
\newblock Weak-to-strong generalization: Eliciting strong capabilities with
  weak supervision.
\newblock \emph{ArXiv}, abs/2312.09390, 2023.
\newblock URL \url{https://api.semanticscholar.org/CorpusID:266312608}.

\bibitem[Chen et~al.(2021)Chen, Tworek, Jun, Yuan, Pinto, Kaplan, Edwards,
  Burda, Joseph, Brockman, et~al.]{chen2021evaluating}
Mark Chen, Jerry Tworek, Heewoo Jun, Qiming Yuan, Henrique Ponde de~Oliveira
  Pinto, Jared Kaplan, Harri Edwards, Yuri Burda, Nicholas Joseph, Greg
  Brockman, et~al.
\newblock Evaluating large language models trained on code.
\newblock \emph{arXiv preprint arXiv:2107.03374}, 2021.

\bibitem[Chiang et~al.(2023)Chiang, Li, Lin, Sheng, Wu, Zhang, Zheng, Zhuang,
  Zhuang, Gonzalez, Stoica, and Xing]{vicuna2023}
Wei-Lin Chiang, Zhuohan Li, Zi~Lin, Ying Sheng, Zhanghao Wu, Hao Zhang, Lianmin
  Zheng, Siyuan Zhuang, Yonghao Zhuang, Joseph~E. Gonzalez, Ion Stoica, and
  Eric~P. Xing.
\newblock Vicuna: An open-source chatbot impressing gpt-4 with 90\%* chatgpt
  quality, March 2023.
\newblock URL \url{https://lmsys.org/blog/2023-03-30-vicuna/}.

\bibitem[Christiano et~al.(2018)Christiano, Shlegeris, and
  Amodei]{Christiano2018SupervisingSL}
Paul~Francis Christiano, Buck Shlegeris, and Dario Amodei.
\newblock Supervising strong learners by amplifying weak experts.
\newblock \emph{ArXiv}, abs/1810.08575, 2018.
\newblock URL \url{https://api.semanticscholar.org/CorpusID:53041432}.

\bibitem[Clark et~al.(2019)Clark, Lee, Chang, Kwiatkowski, Collins, and
  Toutanova]{clark2019boolq}
Christopher Clark, Kenton Lee, Ming-Wei Chang, Tom Kwiatkowski, Michael
  Collins, and Kristina Toutanova.
\newblock Boolq: Exploring the surprising difficulty of natural yes/no
  questions.
\newblock \emph{arXiv preprint arXiv:1905.10044}, 2019.

\bibitem[Clark et~al.(2018)Clark, Cowhey, Etzioni, Khot, Sabharwal, Schoenick,
  and Tafjord]{clark2018think}
Peter Clark, Isaac Cowhey, Oren Etzioni, Tushar Khot, Ashish Sabharwal, Carissa
  Schoenick, and Oyvind Tafjord.
\newblock Think you have solved question answering? try arc, the ai2 reasoning
  challenge.
\newblock \emph{arXiv preprint arXiv:1803.05457}, 2018.

\bibitem[Cobbe et~al.(2021)Cobbe, Kosaraju, Bavarian, Chen, Jun, Kaiser,
  Plappert, Tworek, Hilton, Nakano, et~al.]{cobbe2021training}
Karl Cobbe, Vineet Kosaraju, Mohammad Bavarian, Mark Chen, Heewoo Jun, Lukasz
  Kaiser, Matthias Plappert, Jerry Tworek, Jacob Hilton, Reiichiro Nakano,
  et~al.
\newblock Training verifiers to solve math word problems.
\newblock \emph{arXiv preprint arXiv:2110.14168}, 2021.

\bibitem[Ding et~al.(2023)Ding, Chen, Xu, Qin, Zheng, Hu, Liu, Sun, and
  Zhou]{ding2023enhancing}
Ning Ding, Yulin Chen, Bokai Xu, Yujia Qin, Zhi Zheng, Shengding Hu, Zhiyuan
  Liu, Maosong Sun, and Bowen Zhou.
\newblock Enhancing chat language models by scaling high-quality instructional
  conversations.
\newblock \emph{arXiv preprint arXiv:2305.14233}, 2023.

\bibitem[Dubois et~al.(2023)Dubois, Li, Taori, Zhang, Gulrajani, Ba, Guestrin,
  Liang, and Hashimoto]{dubois2023alpacafarm}
Yann Dubois, Xuechen Li, Rohan Taori, Tianyi Zhang, Ishaan Gulrajani, Jimmy Ba,
  Carlos Guestrin, Percy Liang, and Tatsunori~B Hashimoto.
\newblock Alpacafarm: A simulation framework for methods that learn from human
  feedback.
\newblock \emph{arXiv preprint arXiv:2305.14387}, 2023.

\bibitem[Ethayarajh et~al.(2022)Ethayarajh, Choi, and
  Swayamdipta]{pmlr-v162-ethayarajh22a}
Kawin Ethayarajh, Yejin Choi, and Swabha Swayamdipta.
\newblock Understanding dataset difficulty with $\mathcal{V}$-usable
  information.
\newblock In Kamalika Chaudhuri, Stefanie Jegelka, Le~Song, Csaba Szepesvari,
  Gang Niu, and Sivan Sabato (eds.), \emph{Proceedings of the 39th
  International Conference on Machine Learning}, volume 162 of
  \emph{Proceedings of Machine Learning Research}, pp.\  5988--6008. PMLR,
  17--23 Jul 2022.

\bibitem[Gao et~al.(2021)Gao, Tow, Biderman, Black, DiPofi, Foster, Golding,
  Hsu, McDonell, Muennighoff, Phang, Reynolds, Tang, Thite, Wang, Wang, and
  Zou]{eval-harness}
Leo Gao, Jonathan Tow, Stella Biderman, Sid Black, Anthony DiPofi, Charles
  Foster, Laurence Golding, Jeffrey Hsu, Kyle McDonell, Niklas Muennighoff,
  Jason Phang, Laria Reynolds, Eric Tang, Anish Thite, Ben Wang, Kevin Wang,
  and Andy Zou.
\newblock A framework for few-shot language model evaluation, September 2021.
\newblock URL \url{https://doi.org/10.5281/zenodo.5371628}.

\bibitem[Glaese et~al.(2022)Glaese, McAleese, Trkebacz, Aslanides, Firoiu,
  Ewalds, Rauh, Weidinger, Chadwick, Thacker, et~al.]{glaese2022improving}
Amelia Glaese, Nat McAleese, Maja Trkebacz, John Aslanides, Vlad Firoiu, Timo
  Ewalds, Maribeth Rauh, Laura Weidinger, Martin Chadwick, Phoebe Thacker,
  et~al.
\newblock Improving alignment of dialogue agents via targeted human judgements.
\newblock \emph{arXiv preprint arXiv:2209.14375}, 2022.

\bibitem[Hendrycks et~al.(2021)Hendrycks, Burns, Kadavath, Arora, Basart, Tang,
  Song, and Steinhardt]{hendrycks2021measuring}
Dan Hendrycks, Collin Burns, Saurav Kadavath, Akul Arora, Steven Basart, Eric
  Tang, Dawn Song, and Jacob Steinhardt.
\newblock Measuring mathematical problem solving with the math dataset.
\newblock \emph{arXiv preprint arXiv:2103.03874}, 2021.

\bibitem[Hu et~al.(2023)Hu, Luo, Wang, Cheng, Liu, and Sun]{hu2023falseqa}
Shengding Hu, Yifan Luo, Huadong Wang, Xingyi Cheng, Zhiyuan Liu, and Maosong
  Sun.
\newblock Won't get fooled again: Answering questions with false premises.
\newblock In Anna Rogers, Jordan~L. Boyd{-}Graber, and Naoaki Okazaki (eds.),
  \emph{Proceedings of the 61st Annual Meeting of the Association for
  Computational Linguistics (Volume 1: Long Papers), {ACL} 2023, Toronto,
  Canada, July 9-14, 2023}, 2023.

\bibitem[IDEA-CCNL(2021)]{Fengshenbang-LM}
IDEA-CCNL.
\newblock Fengshenbang-lm.
\newblock \url{https://github.com/IDEA-CCNL/Fengshenbang-LM}, 2021.

\bibitem[Irving et~al.(2018)Irving, Christiano, and Amodei]{Irving2018AISV}
Geoffrey Irving, Paul~Francis Christiano, and Dario Amodei.
\newblock Ai safety via debate.
\newblock \emph{ArXiv}, abs/1805.00899, 2018.
\newblock URL \url{https://api.semanticscholar.org/CorpusID:22050710}.

\bibitem[Ji et~al.(2023)Ji, Liu, Dai, Pan, Zhang, Bian, Sun, Wang, and
  Yang]{ji2023beavertails}
Jiaming Ji, Mickel Liu, Juntao Dai, Xuehai Pan, Chi Zhang, Ce~Bian, Ruiyang
  Sun, Yizhou Wang, and Yaodong Yang.
\newblock Beavertails: Towards improved safety alignment of llm via a
  human-preference dataset.
\newblock \emph{arXiv preprint arXiv:2307.04657}, 2023.

\bibitem[Joshi et~al.(2017)Joshi, Choi, Weld, and
  Zettlemoyer]{joshi2017triviaqa}
Mandar Joshi, Eunsol Choi, Daniel~S Weld, and Luke Zettlemoyer.
\newblock Triviaqa: A large scale distantly supervised challenge dataset for
  reading comprehension.
\newblock \emph{arXiv preprint arXiv:1705.03551}, 2017.

\bibitem[Khashabi et~al.(2018)Khashabi, Chaturvedi, Roth, Upadhyay, and
  Roth]{khashabi2018looking}
Daniel Khashabi, Snigdha Chaturvedi, Michael Roth, Shyam Upadhyay, and Dan
  Roth.
\newblock Looking beyond the surface: A challenge set for reading comprehension
  over multiple sentences.
\newblock In \emph{Proceedings of the 2018 Conference of the North American
  Chapter of the Association for Computational Linguistics: Human Language
  Technologies, Volume 1 (Long Papers)}, pp.\  252--262, 2018.

\bibitem[Kwiatkowski et~al.(2019)Kwiatkowski, Palomaki, Redfield, Collins,
  Parikh, Alberti, Epstein, Polosukhin, Devlin, Lee,
  et~al.]{kwiatkowski2019natural}
Tom Kwiatkowski, Jennimaria Palomaki, Olivia Redfield, Michael Collins, Ankur
  Parikh, Chris Alberti, Danielle Epstein, Illia Polosukhin, Jacob Devlin,
  Kenton Lee, et~al.
\newblock Natural questions: a benchmark for question answering research.
\newblock \emph{Transactions of the Association for Computational Linguistics},
  7:\penalty0 453--466, 2019.

\bibitem[Lai et~al.(2017)Lai, Xie, Liu, Yang, and Hovy]{lai2017race}
Guokun Lai, Qizhe Xie, Hanxiao Liu, Yiming Yang, and Eduard Hovy.
\newblock Race: Large-scale reading comprehension dataset from examinations.
\newblock \emph{arXiv preprint arXiv:1704.04683}, 2017.

\bibitem[Lee et~al.(2023)Lee, Phatale, Mansoor, Lu, Mesnard, Bishop, Carbune,
  and Rastogi]{lee2023rlaif}
Harrison Lee, Samrat Phatale, Hassan Mansoor, Kellie Lu, Thomas Mesnard, Colton
  Bishop, Victor Carbune, and Abhinav Rastogi.
\newblock Rlaif: Scaling reinforcement learning from human feedback with ai
  feedback.
\newblock \emph{arXiv preprint arXiv:2309.00267}, 2023.

\bibitem[Leike et~al.(2018)Leike, Krueger, Everitt, Martic, Maini, and
  Legg]{Leike2018ScalableAA}
Jan Leike, David Krueger, Tom Everitt, Miljan Martic, Vishal Maini, and Shane
  Legg.
\newblock Scalable agent alignment via reward modeling: a research direction.
\newblock \emph{ArXiv}, abs/1811.07871, 2018.
\newblock URL \url{https://api.semanticscholar.org/CorpusID:53745764}.

\bibitem[Li et~al.(2023)Li, Zhang, Dubois, Taori, Gulrajani, Guestrin, Liang,
  and Hashimoto]{alpaca_eval}
Xuechen Li, Tianyi Zhang, Yann Dubois, Rohan Taori, Ishaan Gulrajani, Carlos
  Guestrin, Percy Liang, and Tatsunori~B. Hashimoto.
\newblock Alpacaeval: An automatic evaluator of instruction-following models.
\newblock \url{https://github.com/tatsu-lab/alpaca_eval}, 2023.

\bibitem[Lightman et~al.(2023)Lightman, Kosaraju, Burda, Edwards, Baker, Lee,
  Leike, Schulman, Sutskever, and Cobbe]{lightman2023let}
Hunter Lightman, Vineet Kosaraju, Yura Burda, Harri Edwards, Bowen Baker, Teddy
  Lee, Jan Leike, John Schulman, Ilya Sutskever, and Karl Cobbe.
\newblock Let's verify step by step.
\newblock \emph{arXiv preprint arXiv:2305.20050}, 2023.

\bibitem[Lin et~al.(2022)Lin, Hilton, and Evans]{lin2022truthfulqa}
Stephanie Lin, Jacob Hilton, and Owain Evans.
\newblock Truthfulqa: Measuring how models mimic human falsehoods.
\newblock In Smaranda Muresan, Preslav Nakov, and Aline Villavicencio (eds.),
  \emph{Proceedings of the 60th Annual Meeting of the Association for
  Computational Linguistics (Volume 1: Long Papers), {ACL} 2022, Dublin,
  Ireland, May 22-27, 2022}, 2022.

\bibitem[Longpre et~al.(2023)Longpre, Hou, Vu, Webson, Chung, Tay, Zhou, Le,
  Zoph, Wei, and Roberts]{longpre2023flan}
Shayne Longpre, Le~Hou, Tu~Vu, Albert Webson, Hyung~Won Chung, Yi~Tay, Denny
  Zhou, Quoc~V. Le, Barret Zoph, Jason Wei, and Adam Roberts.
\newblock The flan collection: Designing data and methods for effective
  instruction tuning.
\newblock \emph{CoRR}, abs/2301.13688, 2023.

\bibitem[Mihaylov et~al.(2018)Mihaylov, Clark, Khot, and
  Sabharwal]{mihaylov2018can}
Todor Mihaylov, Peter Clark, Tushar Khot, and Ashish Sabharwal.
\newblock Can a suit of armor conduct electricity? a new dataset for open book
  question answering.
\newblock \emph{arXiv preprint arXiv:1809.02789}, 2018.

\bibitem[MosaicML(2023)]{MosaicML2023Introducing}
MosaicML.
\newblock Introducing mpt-30b: Raising the bar for open-source foundation
  models, 2023.
\newblock URL \url{www.mosaicml.com/blog/mpt-30b}.
\newblock Accessed: 2023-06-22.

\bibitem[Mukherjee et~al.(2023)Mukherjee, Mitra, Jawahar, Agarwal, Palangi, and
  Awadallah]{mukherjee2023orca}
Subhabrata Mukherjee, Arindam Mitra, Ganesh Jawahar, Sahaj Agarwal, Hamid
  Palangi, and Ahmed~Hassan Awadallah.
\newblock Orca: Progressive learning from complex explanation traces of
  {GPT-4}.
\newblock \emph{CoRR}, abs/2306.02707, 2023.

\bibitem[Nakano et~al.(2021)Nakano, Hilton, Balaji, Wu, Ouyang, Kim, Hesse,
  Jain, Kosaraju, Saunders, et~al.]{nakano2021webgpt}
Reiichiro Nakano, Jacob Hilton, Suchir Balaji, Jeff Wu, Long Ouyang, Christina
  Kim, Christopher Hesse, Shantanu Jain, Vineet Kosaraju, William Saunders,
  et~al.
\newblock Webgpt: Browser-assisted question-answering with human feedback.
\newblock \emph{arXiv preprint arXiv:2112.09332}, 2021.

\bibitem[OpenAI(2022)]{openai2022chatgpt}
OpenAI.
\newblock Chatgpt: Optimizing language models for dialogue, 2022.

\bibitem[OpenAI(2023)]{openai2023gpt4}
OpenAI.
\newblock Gpt-4 technical report, 2023.

\bibitem[Ouyang et~al.(2022)Ouyang, Wu, Jiang, Almeida, Wainwright, Mishkin,
  Zhang, Agarwal, Slama, Ray, et~al.]{ouyang2022training}
Long Ouyang, Jeffrey Wu, Xu~Jiang, Diogo Almeida, Carroll Wainwright, Pamela
  Mishkin, Chong Zhang, Sandhini Agarwal, Katarina Slama, Alex Ray, et~al.
\newblock Training language models to follow instructions with human feedback.
\newblock \emph{Advances in Neural Information Processing Systems},
  35:\penalty0 27730--27744, 2022.

\bibitem[Park et~al.(2023)Park, O'Brien, Cai, Morris, Liang, and
  Bernstein]{park2023generative}
Joon~Sung Park, Joseph~C O'Brien, Carrie~J Cai, Meredith~Ringel Morris, Percy
  Liang, and Michael~S Bernstein.
\newblock Generative agents: Interactive simulacra of human behavior.
\newblock \emph{arXiv preprint arXiv:2304.03442}, 2023.

\bibitem[Perez et~al.(2022)Perez, Huang, Song, Cai, Ring, Aslanides, Glaese,
  McAleese, and Irving]{perez2022red}
Ethan Perez, Saffron Huang, H.~Francis Song, Trevor Cai, Roman Ring, John
  Aslanides, Amelia Glaese, Nat McAleese, and Geoffrey Irving.
\newblock Red teaming language models with language models.
\newblock In \emph{Proceedings of the 2022 Conference on Empirical Methods in
  Natural Language Processing, {EMNLP} 2022, Abu Dhabi, United Arab Emirates,
  December 7-11, 2022}, pp.\  3419--3448. Association for Computational
  Linguistics, 2022.
\newblock \doi{10.18653/v1/2022.emnlp-main.225}.
\newblock URL \url{https://doi.org/10.18653/v1/2022.emnlp-main.225}.

\bibitem[Qian et~al.(2023)Qian, Cong, Yang, Chen, Su, Xu, Liu, and
  Sun]{qian2023communicative}
Chen Qian, Xin Cong, Cheng Yang, Weize Chen, Yusheng Su, Juyuan Xu, Zhiyuan
  Liu, and Maosong Sun.
\newblock Communicative agents for software development.
\newblock \emph{arXiv preprint arXiv:2307.07924}, 2023.

\bibitem[Qin et~al.(2023)Qin, Hu, Lin, Chen, Ding, Cui, Zeng, Huang, Xiao, Han,
  et~al.]{qin2023tool}
Yujia Qin, Shengding Hu, Yankai Lin, Weize Chen, Ning Ding, Ganqu Cui, Zheni
  Zeng, Yufei Huang, Chaojun Xiao, Chi Han, et~al.
\newblock Tool learning with foundation models.
\newblock \emph{arXiv preprint arXiv:2304.08354}, 2023.

\bibitem[Rafailov et~al.(2023)Rafailov, Sharma, Mitchell, Ermon, Manning, and
  Finn]{rafailov2023direct}
Rafael Rafailov, Archit Sharma, Eric Mitchell, Stefano Ermon, Christopher~D
  Manning, and Chelsea Finn.
\newblock Direct preference optimization: Your language model is secretly a
  reward model.
\newblock \emph{arXiv preprint arXiv:2305.18290}, 2023.

\bibitem[Sanh et~al.(2022)Sanh, Webson, Raffel, Bach, Sutawika, Alyafeai,
  Chaffin, Stiegler, Raja, Dey, Bari, Xu, Thakker, Sharma, Szczechla, Kim,
  Chhablani, Nayak, Datta, Chang, Jiang, Wang, Manica, Shen, Yong, Pandey,
  Bawden, Wang, Neeraj, Rozen, Sharma, Santilli, F{\'{e}}vry, Fries, Teehan,
  Scao, Biderman, Gao, Wolf, and Rush]{sanh2022multitask}
Victor Sanh, Albert Webson, Colin Raffel, Stephen~H. Bach, Lintang Sutawika,
  Zaid Alyafeai, Antoine Chaffin, Arnaud Stiegler, Arun Raja, Manan Dey,
  M~Saiful Bari, Canwen Xu, Urmish Thakker, Shanya~Sharma Sharma, Eliza
  Szczechla, Taewoon Kim, Gunjan Chhablani, Nihal~V. Nayak, Debajyoti Datta,
  Jonathan Chang, Mike~Tian{-}Jian Jiang, Han Wang, Matteo Manica, Sheng Shen,
  Zheng~Xin Yong, Harshit Pandey, Rachel Bawden, Thomas Wang, Trishala Neeraj,
  Jos Rozen, Abheesht Sharma, Andrea Santilli, Thibault F{\'{e}}vry, Jason~Alan
  Fries, Ryan Teehan, Teven~Le Scao, Stella Biderman, Leo Gao, Thomas Wolf, and
  Alexander~M. Rush.
\newblock Multitask prompted training enables zero-shot task generalization.
\newblock In \emph{The Tenth International Conference on Learning
  Representations, {ICLR} 2022, Virtual Event, April 25-29, 2022}, 2022.
\newblock URL \url{https://openreview.net/forum?id=9Vrb9D0WI4}.

\bibitem[Saunders et~al.(2022)Saunders, Yeh, Wu, Bills, Ouyang, Ward, and
  Leike]{saunders2022self}
William Saunders, Catherine Yeh, Jeff Wu, Steven Bills, Long Ouyang, Jonathan
  Ward, and Jan Leike.
\newblock Self-critiquing models for assisting human evaluators.
\newblock \emph{CoRR}, abs/2206.05802, 2022.
\newblock \doi{10.48550/ARXIV.2206.05802}.
\newblock URL \url{https://doi.org/10.48550/arXiv.2206.05802}.

\bibitem[Schulman et~al.(2017)Schulman, Wolski, Dhariwal, Radford, and
  Klimov]{schulman2017proximal}
John Schulman, Filip Wolski, Prafulla Dhariwal, Alec Radford, and Oleg Klimov.
\newblock Proximal policy optimization algorithms.
\newblock \emph{arXiv preprint arXiv:1707.06347}, 2017.

\bibitem[Shinn et~al.(2023)Shinn, Cassano, Labash, Gopinath, Narasimhan, and
  Yao]{shinn2023reflexion}
Noah Shinn, Federico Cassano, Beck Labash, Ashwin Gopinath, Karthik Narasimhan,
  and Shunyu Yao.
\newblock Reflexion: Language agents with verbal reinforcement learning, 2023.

\bibitem[Stiennon et~al.(2020)Stiennon, Ouyang, Wu, Ziegler, Lowe, Voss,
  Radford, Amodei, and Christiano]{stiennon2020learning}
Nisan Stiennon, Long Ouyang, Jeffrey Wu, Daniel Ziegler, Ryan Lowe, Chelsea
  Voss, Alec Radford, Dario Amodei, and Paul~F Christiano.
\newblock Learning to summarize with human feedback.
\newblock \emph{Advances in Neural Information Processing Systems},
  33:\penalty0 3008--3021, 2020.

\bibitem[Sun et~al.(2023)Sun, Shen, Zhou, Zhang, Chen, Cox, Yang, and
  Gan]{Sun2023PrincipleDrivenSO}
Zhiqing Sun, Yikang Shen, Qinhong Zhou, Hongxin Zhang, Zhenfang Chen, David~D.
  Cox, Yiming Yang, and Chuang Gan.
\newblock Principle-driven self-alignment of language models from scratch with
  minimal human supervision.
\newblock \emph{ArXiv}, abs/2305.03047, 2023.

\bibitem[Talmor et~al.(2018)Talmor, Herzig, Lourie, and
  Berant]{talmor2018commonsenseqa}
Alon Talmor, Jonathan Herzig, Nicholas Lourie, and Jonathan Berant.
\newblock Commonsenseqa: A question answering challenge targeting commonsense
  knowledge.
\newblock \emph{arXiv preprint arXiv:1811.00937}, 2018.

\bibitem[Taori et~al.(2023)Taori, Gulrajani, Zhang, Dubois, Li, Guestrin,
  Liang, and Hashimoto]{taori2023stanford}
Rohan Taori, Ishaan Gulrajani, Tianyi Zhang, Yann Dubois, Xuechen Li, Carlos
  Guestrin, Percy Liang, and Tatsunori~B Hashimoto.
\newblock Stanford alpaca: An instruction-following llama model, 2023.

\bibitem[Touvron et~al.(2023{\natexlab{a}})Touvron, Lavril, Izacard, Martinet,
  Lachaux, Lacroix, Rozi{\`e}re, Goyal, Hambro, Azhar,
  et~al.]{touvron2023llama1}
Hugo Touvron, Thibaut Lavril, Gautier Izacard, Xavier Martinet, Marie-Anne
  Lachaux, Timoth{\'e}e Lacroix, Baptiste Rozi{\`e}re, Naman Goyal, Eric
  Hambro, Faisal Azhar, et~al.
\newblock Llama: Open and efficient foundation language models.
\newblock \emph{arXiv preprint arXiv:2302.13971}, 2023{\natexlab{a}}.

\bibitem[Touvron et~al.(2023{\natexlab{b}})Touvron, Martin, Stone, Albert,
  Almahairi, Babaei, Bashlykov, Batra, Bhargava, Bhosale,
  et~al.]{touvron2023llama}
Hugo Touvron, Louis Martin, Kevin Stone, Peter Albert, Amjad Almahairi, Yasmine
  Babaei, Nikolay Bashlykov, Soumya Batra, Prajjwal Bhargava, Shruti Bhosale,
  et~al.
\newblock Llama 2: Open foundation and fine-tuned chat models.
\newblock \emph{arXiv preprint arXiv:2307.09288}, 2023{\natexlab{b}}.

\bibitem[Tunstall et~al.(2023)Tunstall, Lambert, Rajani, Beeching, Le~Scao, von
  Werra, Han, Schmid, and Rush]{Tunstall2023starchat-alpha}
Lewis Tunstall, Nathan Lambert, Nazneen Rajani, Edward Beeching, Teven Le~Scao,
  Leandro von Werra, Sheon Han, Philipp Schmid, and Alexander Rush.
\newblock Creating a coding assistant with starcoder.
\newblock \emph{Hugging Face Blog}, 2023.
\newblock https://huggingface.co/blog/starchat.

\bibitem[Wang et~al.(2023{\natexlab{a}})Wang, Cheng, Zhan, Li, Song, and
  Liu]{wang2023openchat}
Guan Wang, Sijie Cheng, Xianyuan Zhan, Xiangang Li, Sen Song, and Yang Liu.
\newblock Openchat: Advancing open-source language models with mixed-quality
  data.
\newblock \emph{arXiv preprint arXiv:2309.11235}, 2023{\natexlab{a}}.

\bibitem[Wang et~al.(2023{\natexlab{b}})Wang, Li, Chen, Cai, Zhu, Lin, Cao,
  Liu, Liu, and Sui]{wang2023large}
Peiyi Wang, Lei Li, Liang Chen, Zefan Cai, Dawei Zhu, Binghuai Lin, Yunbo Cao,
  Qi~Liu, Tianyu Liu, and Zhifang Sui.
\newblock Large language models are not fair evaluators, 2023{\natexlab{b}}.

\bibitem[Wang et~al.(2023{\natexlab{c}})Wang, Yu, Tan, O'Brien, Pasunuru,
  Dwivedi-Yu, Golovneva, Zettlemoyer, Fazel-Zarandi, and
  Celikyilmaz]{wang2023shepherd}
Tianlu Wang, Ping Yu, Xiaoqing~Ellen Tan, Sean O'Brien, Ramakanth Pasunuru,
  Jane Dwivedi-Yu, Olga Golovneva, Luke Zettlemoyer, Maryam Fazel-Zarandi, and
  Asli Celikyilmaz.
\newblock Shepherd: A critic for language model generation.
\newblock \emph{arXiv preprint arXiv:2308.04592}, 2023{\natexlab{c}}.

\bibitem[Wang et~al.(2023{\natexlab{d}})Wang, Wang, Liu, Chen, Yuan, Peng, and
  Ji]{wang2023mint}
Xingyao Wang, Zihan Wang, Jiateng Liu, Yangyi Chen, Lifan Yuan, Hao Peng, and
  Heng Ji.
\newblock Mint: Evaluating llms in multi-turn interaction with tools and
  language feedback.
\newblock \emph{arXiv preprint arXiv:2309.10691}, 2023{\natexlab{d}}.

\bibitem[Wang et~al.(2022)Wang, Mishra, Alipoormolabashi, Kordi, Mirzaei, Naik,
  Ashok, Dhanasekaran, Arunkumar, Stap, Pathak, Karamanolakis, Lai, Purohit,
  Mondal, Anderson, Kuznia, Doshi, Pal, Patel, Moradshahi, Parmar, Purohit,
  Varshney, Kaza, Verma, Puri, Karia, Doshi, Sampat, Mishra, A, Patro, Dixit,
  and Shen]{wang2022super}
Yizhong Wang, Swaroop Mishra, Pegah Alipoormolabashi, Yeganeh Kordi, Amirreza
  Mirzaei, Atharva Naik, Arjun Ashok, Arut~Selvan Dhanasekaran, Anjana
  Arunkumar, David Stap, Eshaan Pathak, Giannis Karamanolakis, Haizhi~Gary Lai,
  Ishan Purohit, Ishani Mondal, Jacob Anderson, Kirby Kuznia, Krima Doshi,
  Kuntal~Kumar Pal, Maitreya Patel, Mehrad Moradshahi, Mihir Parmar, Mirali
  Purohit, Neeraj Varshney, Phani~Rohitha Kaza, Pulkit Verma, Ravsehaj~Singh
  Puri, Rushang Karia, Savan Doshi, Shailaja~Keyur Sampat, Siddhartha Mishra,
  Sujan~Reddy A, Sumanta Patro, Tanay Dixit, and Xudong Shen.
\newblock Super-naturalinstructions: Generalization via declarative
  instructions on 1600+ {NLP} tasks.
\newblock In \emph{Proceedings of the 2022 Conference on Empirical Methods in
  Natural Language Processing, {EMNLP} 2022, Abu Dhabi, United Arab Emirates,
  December 7-11, 2022}, pp.\  5085--5109, 2022.
\newblock URL \url{https://doi.org/10.18653/v1/2022.emnlp-main.340}.

\bibitem[Wei et~al.(2022{\natexlab{a}})Wei, Bosma, Zhao, Guu, Yu, Lester, Du,
  Dai, and Le]{wei2022finetuned}
Jason Wei, Maarten Bosma, Vincent~Y. Zhao, Kelvin Guu, Adams~Wei Yu, Brian
  Lester, Nan Du, Andrew~M. Dai, and Quoc~V. Le.
\newblock Finetuned language models are zero-shot learners.
\newblock In \emph{The Tenth International Conference on Learning
  Representations, {ICLR} 2022, Virtual Event, April 25-29, 2022},
  2022{\natexlab{a}}.
\newblock URL \url{https://openreview.net/forum?id=gEZrGCozdqR}.

\bibitem[Wei et~al.(2022{\natexlab{b}})Wei, Wang, Schuurmans, Bosma, Xia, Chi,
  Le, Zhou, et~al.]{wei2022chain}
Jason Wei, Xuezhi Wang, Dale Schuurmans, Maarten Bosma, Fei Xia, Ed~Chi, Quoc~V
  Le, Denny Zhou, et~al.
\newblock Chain-of-thought prompting elicits reasoning in large language
  models.
\newblock \emph{Advances in Neural Information Processing Systems},
  35:\penalty0 24824--24837, 2022{\natexlab{b}}.

\bibitem[Wu et~al.(2023)Wu, Hu, Shi, Dziri, Suhr, Ammanabrolu, Smith,
  Ostendorf, and Hajishirzi]{wu2023fine}
Zeqiu Wu, Yushi Hu, Weijia Shi, Nouha Dziri, Alane Suhr, Prithviraj
  Ammanabrolu, Noah~A Smith, Mari Ostendorf, and Hannaneh Hajishirzi.
\newblock Fine-grained human feedback gives better rewards for language model
  training.
\newblock \emph{arXiv preprint arXiv:2306.01693}, 2023.

\bibitem[Xu et~al.(2023)Xu, Sun, Zheng, Geng, Zhao, Feng, Tao, and
  Jiang]{xu2023wizardlm}
Can Xu, Qingfeng Sun, Kai Zheng, Xiubo Geng, Pu~Zhao, Jiazhan Feng, Chongyang
  Tao, and Daxin Jiang.
\newblock Wizardlm: Empowering large language models to follow complex
  instructions.
\newblock \emph{arXiv preprint arXiv:2304.12244}, 2023.

\bibitem[Yao et~al.(2023)Yao, Heinecke, Niebles, Liu, Feng, Xue, Murthy, Chen,
  Zhang, Arpit, et~al.]{yao2023retroformer}
Weiran Yao, Shelby Heinecke, Juan~Carlos Niebles, Zhiwei Liu, Yihao Feng,
  Le~Xue, Rithesh Murthy, Zeyuan Chen, Jianguo Zhang, Devansh Arpit, et~al.
\newblock Retroformer: Retrospective large language agents with policy gradient
  optimization.
\newblock \emph{arXiv preprint arXiv:2308.02151}, 2023.

\bibitem[Ye et~al.(2023)Ye, Jo, Kim, Kim, Hwang, and Seo]{selfee2023}
Seonghyeon Ye, Yongrae Jo, Doyoung Kim, Sungdong Kim, Hyeonbin Hwang, and
  Minjoon Seo.
\newblock Selfee: Iterative self-revising llm empowered by self-feedback
  generation.
\newblock Blog post, May 2023.
\newblock URL \url{https://kaistai.github.io/SelFee/}.

\bibitem[Zellers et~al.(2019)Zellers, Holtzman, Bisk, Farhadi, and
  Choi]{zellers2019hellaswag}
Rowan Zellers, Ari Holtzman, Yonatan Bisk, Ali Farhadi, and Yejin Choi.
\newblock Hellaswag: Can a machine really finish your sentence?
\newblock \emph{arXiv preprint arXiv:1905.07830}, 2019.

\bibitem[Zheng et~al.(2023{\natexlab{a}})Zheng, Chiang, Sheng, Zhuang, Wu,
  Zhuang, Lin, Li, Li, Xing, Zhang, Gonzalez, and Stoica]{zheng2023judging}
Lianmin Zheng, Wei-Lin Chiang, Ying Sheng, Siyuan Zhuang, Zhanghao Wu, Yonghao
  Zhuang, Zi~Lin, Zhuohan Li, Dacheng Li, Eric.~P Xing, Hao Zhang, Joseph~E.
  Gonzalez, and Ion Stoica.
\newblock Judging llm-as-a-judge with mt-bench and chatbot arena.
\newblock \emph{arXiv preprint arXiv:2306.05685}, 2023{\natexlab{a}}.

\bibitem[Zheng et~al.(2023{\natexlab{b}})Zheng, Dou, Gao, Shen, Wang, Liu, Jin,
  Liu, Xiong, Chen, et~al.]{zheng2023secrets}
Rui Zheng, Shihan Dou, Songyang Gao, Wei Shen, Binghai Wang, Yan Liu, Senjie
  Jin, Qin Liu, Limao Xiong, Lu~Chen, et~al.
\newblock Secrets of rlhf in large language models part i: Ppo.
\newblock \emph{arXiv preprint arXiv:2307.04964}, 2023{\natexlab{b}}.

\end{thebibliography}
\bibliographystyle{iclr2024_conference}

% \newpage
\appendix
\onecolumn

\section{Limitations}
In constructing \textsc{UltraFeedback}, we made an assumption that powerful LLMs like GPT-4 are capable of imitating human annotators and fair evaluators. Although more and more works accepted this assumption and demonstrated high agreement between human and LLM feedbacks~\citep{dubois2023alpacafarm, lee2023rlaif, bai2022constitutional}, LLMs still cannot model human preference precisely under all situations. How to efficiently and accurately collect preference data and conduct rigorous evaluation are still challenging. We leave this as future work for further investigation.
Another limitation is that \textsc{UltraFeedback} only provides single-turn dialogues to improve the utility of LLMs due to time and budget restrictions. We will also expand \textsc{UltraFeedback} to cover more tasks and scenarios.

\section{Data Contamination}
\label{sec:data_contamination}
To avoid data contamination which could result in unfair even wrong evaluations, we did careful decontamination for \textsc{UltraFeedback}. Following GPT-3~\citep{brown2020language} and evaluation-harness~\citep{eval-harness}, we search for 13-gram matches between AlpacaEval, Evol-Instruct, and UltraChat test set. We found in total 48 contamination samples and filtered out them. However, we did not conduct a thorough examination of contamination over other evaluation datasets because of the huge amount of datasets. Therefore, we suggest researchers decontaminate \textsc{UltraFeedback} with their evaluation datasets before using it.

\section{UltraFeedback Statistics}
\begin{figure}
    \centering
% \begin{subfigure}{\textwidth}
%     \includegraphics[width=\textwidth]{figures/instruction_following_plot.pdf}
% \end{subfigure}
% \hfill
% \begin{subfigure}{\textwidth}
%     \includegraphics[width=\textwidth]{figures/honesty_plot.pdf}
% \end{subfigure}
% \hfill
% \begin{subfigure}{\textwidth}
%     \includegraphics[width=\textwidth]{figures/truthfulness_plot.pdf}
% \end{subfigure}
% \hfill
% \begin{subfigure}{\textwidth}
%     \includegraphics[width=\textwidth]{figures/helpfulness_plot.pdf}
% \end{subfigure}
% \hfill
\includegraphics[width=0.8\textwidth]{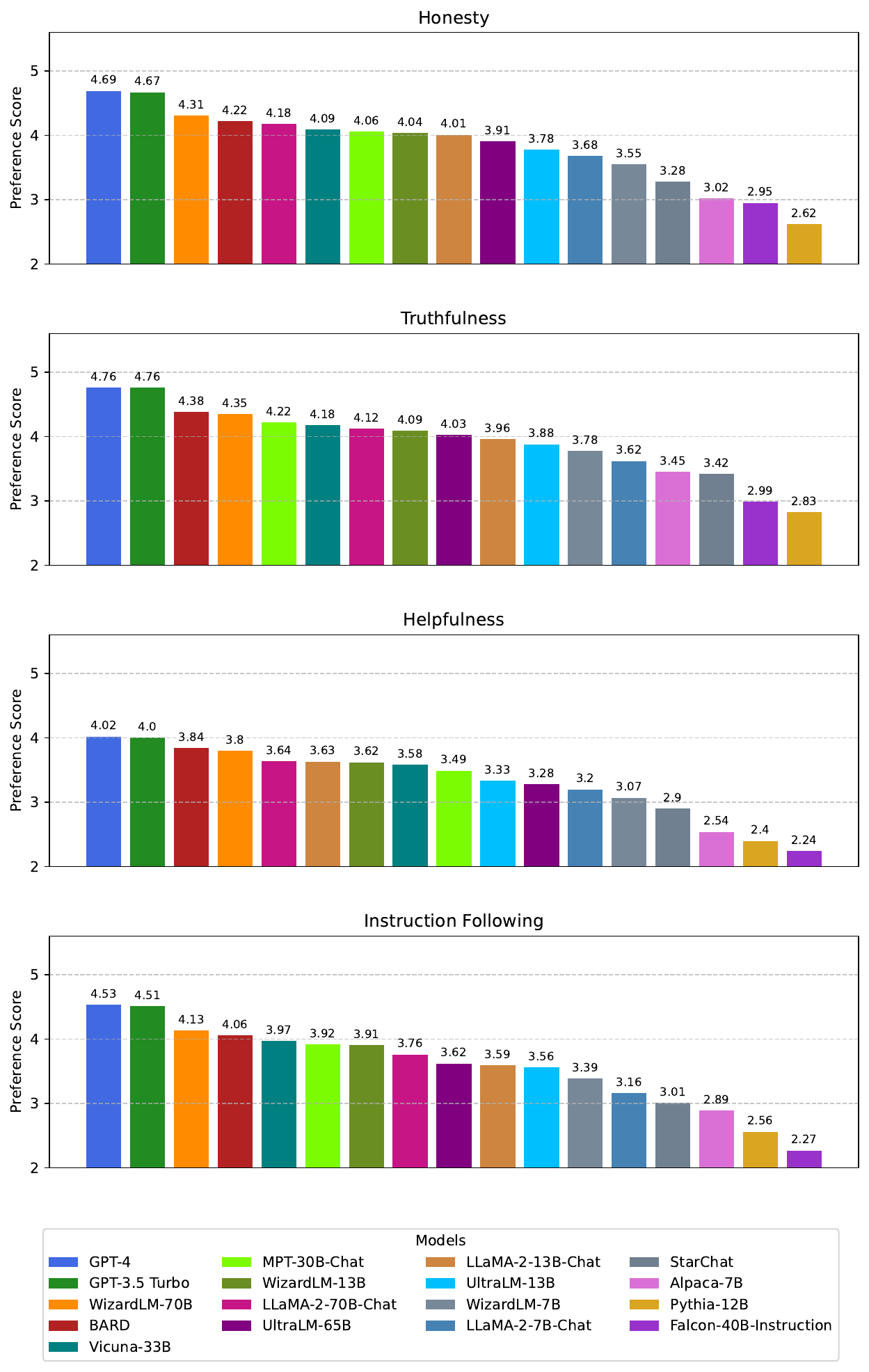}
\caption{Average scores for each model over the four aspects.}
\label{fig:stat}
\end{figure}
We summarize the scores for each model over different aspects in Figure~\ref{fig:stat}. Overall, the rankings are consistent with model capabilities. For example, the GPT series is the best in all aspects, and larger models are generally better than smaller ones. The distinction among different aspects also exists. For instance, the LLaMA2-Chat models received higher scores on honesty, since they are aligned with human values with RLHF~\citep{touvron2023llama}.

\begin{figure}[thb]
    \centering
    \begin{minipage}{0.45\linewidth}
        \centering
        \includegraphics[width=\linewidth]{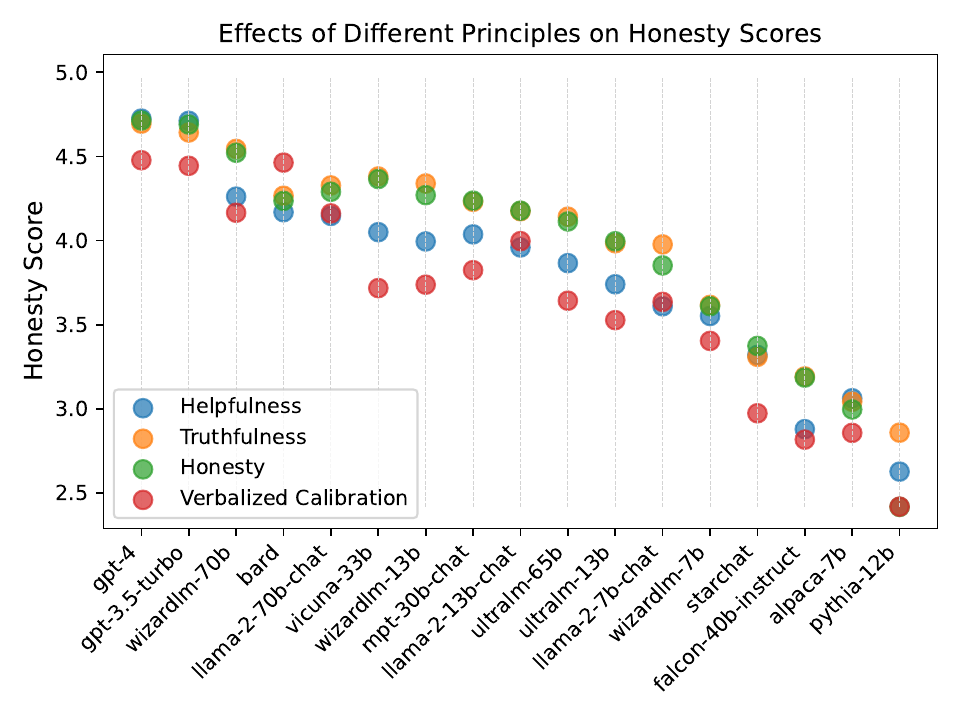}
    \end{minipage}
    \hfill
    \begin{minipage}{0.45\linewidth}
        \centering
        \includegraphics[width=\linewidth]{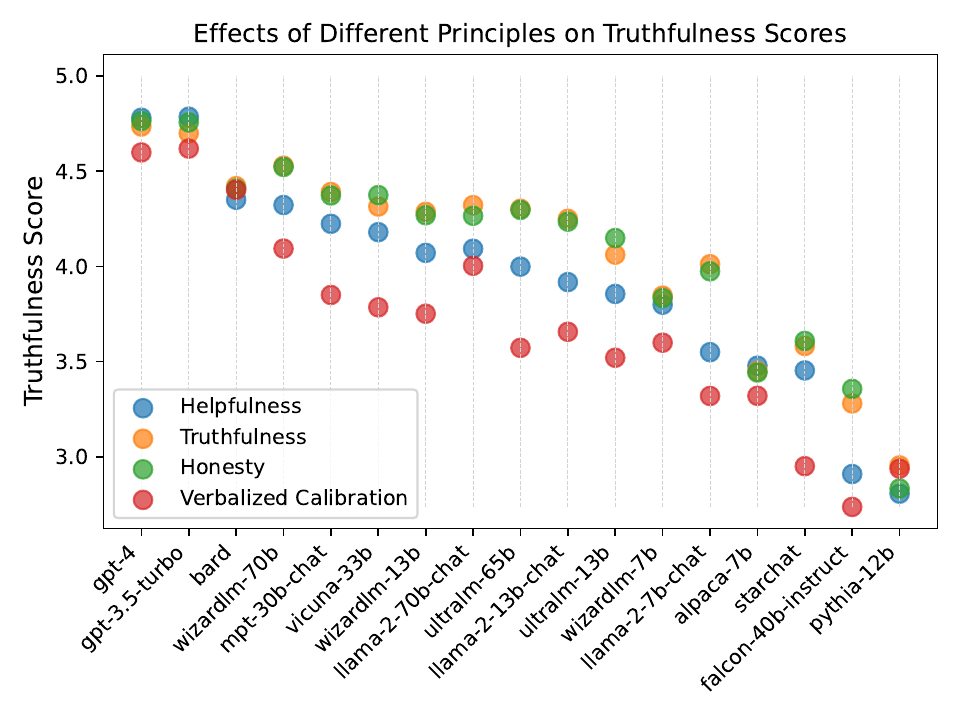}
    \end{minipage}

    \begin{minipage}{0.45\linewidth}
        \centering
        \includegraphics[width=\linewidth]{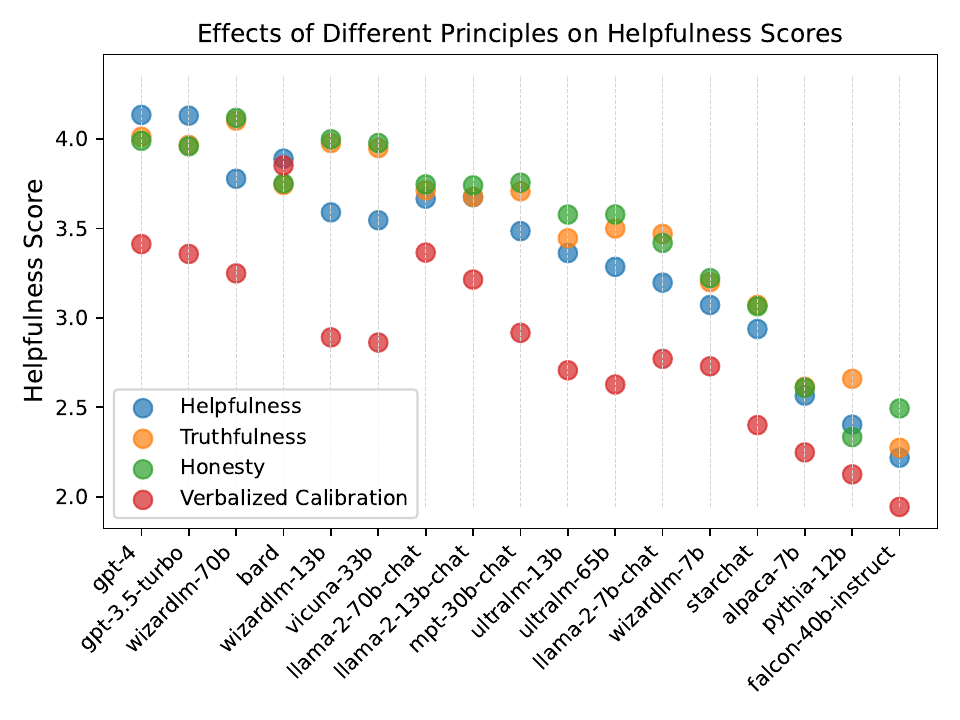}
    \end{minipage}
    \hfill
    \begin{minipage}{0.45\linewidth}
        \centering
        \includegraphics[width=\linewidth]{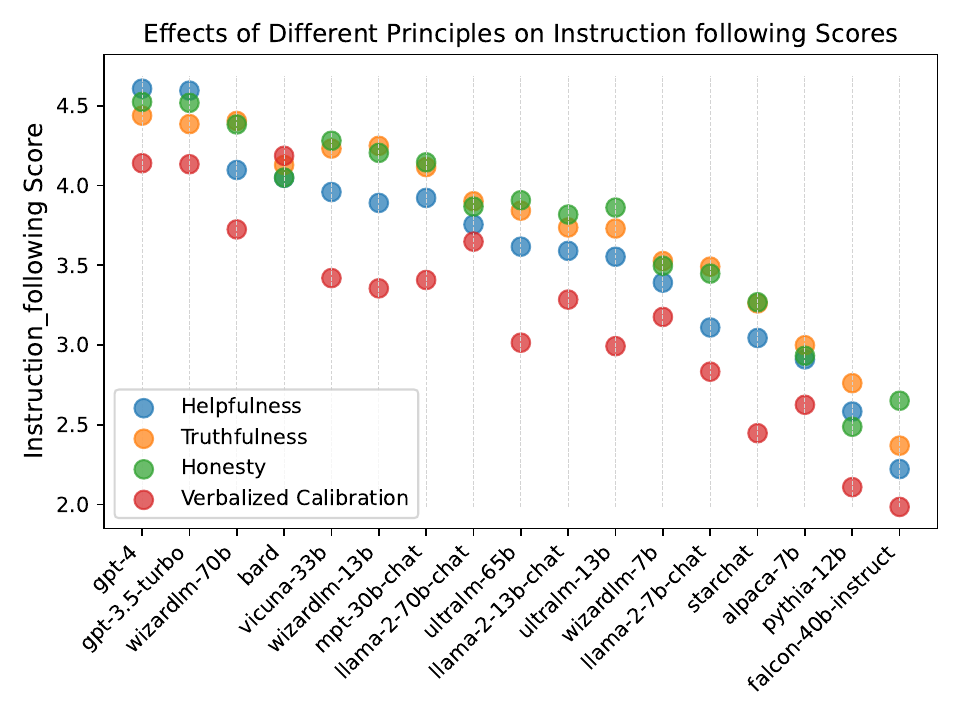}
    \end{minipage}

\caption{Different principles stimulate diverse model behaviors.}
\label{fig:principle_effects}
\end{figure}

We also showcase how different principles stimulate diverse model behaviors. We average the score of each aspect when applying different principles to models, and plot them in Figure \ref{fig:principle_effects}.

\section{Training Details}
\label{sec:implementation_detals}
\subsection{UltraRM}
We construct each comparison pair as a binary selection, with one completion being chosen and the other rejected. We optimize the reward model to select preferred completion by minimizing the binary ranking loss:
\begin{equation}
    \mathcal{L}_{\text {ranking }}=-\log \left(\sigma\left(r_{\theta}\left(x, y_{c}\right)-r_{\theta}\left(x, y_{r}\right)-m(r)\right)\right)
\end{equation}
where $\theta$ represents the reward model, $r_{\theta}\left(x, y_{c}\right)$ is its scalar reward prediction towards the chosen text, $r_{\theta}\left(x, y_{r}\right)$ is that towards the rejected text, and $m\left(r\right)$ is the absolute difference between the annotated reward of two texts. We set the $m\left(r\right)=0$ for datasets with only preference rankings and normalize the margins to $\left(0, 1 \right]$ to avoid training instability due to a mismatch among the score scales of the datasets.
Following \citet{touvron2023llama}, we train the 13B reward model for one epoch with the batch size being 512 pairs (i.e., 1024 completions) and the learning rate being 1e-5. We adopt the cosine learning rate decay strategy with a warm-up ratio of 3\% and a final learning rate of 1e-6.

\subsection{UltraCM}
We train LLaMA2-13B for two epochs with a batch size of 256 and a learning rate of 2e-5. We adopt the same learning rate scheduler as reward modeling.

\section{Experiment Details}
% \subsection{Reward Modeling}
\subsection{Dataset Details for UltraRM Training}
\label{sec:app_data}
We mix \textsc{UltraFeedback}  with other open-source preference datasets for reward modeling.
Stanford SHP is a community-based preference dataset collected from 18 different topics, adopting a strict filtering strategy to ensure text quality and reliability of preferences. We follow the guidelines in the official repository to further filter the dataset, only retaining preferences with a score ratio greater than 2 and using at most 5 comparison pairs for each post via random sampling.
OpenAI Summarize consists of human-written completions and human-annotated preferences, with the instructions being much longer than \textsc{UltraFeedback}. Hence, we include the high-quality dataset to enhance the subsequent reward model for long-text scenarios. We adopt the same comparison pair filtering method to avoid the reward model overfitting certain instructions.
Anthropic Helpful is another human-annotated dataset. We incorporate all its samples into our training dataset to supplement multiturn dialogs data. 
For \textsc{UltraFeedback}, we directly adopt the overall score obtained in critique annotation as the preference score for UltraRM-Overall, while for fine-grained versions, we average the scores of all aspects for each sample as the final preference score.
Finally, the training dataset for our reward model contains a total of $749,702$ comparison pairs, with $340,025$ from \textsc{UltraFeedback}, $198,556$ from Stanford SHP, $92,858$ from OpenAI Summarize, and $118,263$ from Anthropic Helpful.
% \subsubsection{Evaluation Details}
% We note that the SteamSHP model is trained on FLAN-T5-XL \citep{wei2022finetuned}. It inputs preference scores to a pair of completions simultaneously

\subsection{Additional Reward Modeling Experiments}
We observed that the SteamSHP model is different from other reward models in the input format, for it accepts two responses simultaneously and outputs which one is better (text-to-text format). During the experiment, we found that there is a position bias issue for this approach, where the reward model tends to prefer the first responses. To eliminate this issue, we average the scores from two runs exchanging response orders to get the final scores. We report the detailed results in Table \ref{tab:shp_bias}.

\begin{table}[]
\centering
\caption{Reward modeling results for SteamSHP with different sample orders.}
\begin{tabular}{@{}l|cccc@{}}
\toprule
\textbf{Dataset}        & \textbf{\begin{tabular}[c]{@{}c@{}}Anthropic\\      Helpful\end{tabular}} & \textbf{\begin{tabular}[c]{@{}c@{}}OpenAI\\      WebGPT\end{tabular}} & \textbf{\begin{tabular}[c]{@{}c@{}}OpenAI \\      Summ.\end{tabular}} & \textbf{\begin{tabular}[c]{@{}c@{}}Stanford\\      SHP\end{tabular}} \\ \midrule
\textbf{Chosen first}   & 72.0                       & 72.4            & 52.8                  & 71.8                  \\
\textbf{Rejected first} & 38.8                       & 52.9            & 44.0                  & 31.4                  \\
\midrule
\textbf{Avg.}           & 55.4                       & 62.6            & 48.4                  & 51.6                  \\ \bottomrule
\end{tabular}
\label{tab:shp_bias}
\end{table}

\subsection{Critique Modeling}
\label{sec:cm}
\paragraph{Setup.}
To assess the ability of UltraCM to provide reliable critique, we employ GPT-4 to score the quality of critique based on detailed documentation.
we follow \cite{wang2023shepherd} to randomly sample 50 instructions from PIQA~\citep{bisk2020piqa}, OpenBookQA (OBQA)~\citep{mihaylov2018can}, CommonsenseQA~\citep{talmor2018commonsenseqa}, AlpacaFarm~\citep{dubois2023alpacafarm}, and FairEval \citep{wang2023large}. %We do not include TruthfulQA since its training set is involved in the training data of UltraCM. 
We also supplement HumanEval~\citep{chen2021evaluating}, MBPP~\citep{austin2021program}, MATH~\citep{hendrycks2021measuring}, and GSM8K~\citep{cobbe2021training} to evaluate critique quality on coding and math tasks. 
We then generate model completions for the instructions in the same way as Section \ref{sec:ultra_construct}.
We adopt two categories of models for comparison. First, we compare with four general-purpose models, \texttt{gpt-3.5-turbo}, LLaMA2-13B-Chat, Vicuna-13B-v1.5, and WizardLM-13B-v1.2. Then, we adopt two specifically trained critique models, SelFee\footnote{\url{https://huggingface.co/kaist-ai/selfee-13b}} and Shepherd \citep{wang2023shepherd} \footnote{Note that \citet{wang2023shepherd} did not open source their model weights, so we use the model from the community that has been trained on their data: \url{https://huggingface.co/reciprocate/shepherd-13b}}. We apply the baseline models and UltraCM to provide feedback on model completions respectively. Finally, we rate the quality of the critique from 1 to 7 using GPT-4, 1 being the worst and 7 being the best. The prompt is adapted from \cite{wang2023shepherd}. 

\paragraph{Results.}
The scores of feedback quality are presented in Table \ref{tab:feedback_quality}. Overall, the performances of UltraCM almost approach \texttt{gpt-3.5-turbo} and dramatically surpass other models of both categories. 
To be specific, UltraCM achieves comparable performance with \texttt{gpt-3.5-turbo} on commonsense reasoning and mathematics reasoning. However, on AlpacaFarm and code datasets, UltraCM still exhibits deficiencies. Compared with two critique models, we find that (the community-trained) Shepherd almost always fails to provide high-quality feedback. SelFee achieves the highest average scores after \texttt{gpt-3.5-turbo} and UltraCM, but it dramatically falls short on HumanEval and MATH. 
% compare with LLaMA-2 series models-> 
We highlight the comparison between UltraCM and the other three general-purpose models. All four models are trained from LLaMA2-13B, but UltraCM is the only one trained to provide textual critique rather than enhancing knowledge or reasoning capability. However, the feedback of UltraCM consistently gains higher scores than other models across all tasks and datasets, indicating that criticizing is a learnable task and employing an expert critic is more effective than an expert for downstream tasks in providing feedback. With more powerful backbone models, we believe \textsc{UltraFeedback} will greatly benefit autonomous agents~\citep{park2023generative, qin2023tool, qian2023communicative} and feedback learning~\citep{yao2023retroformer, shinn2023reflexion} research.
%\lifan{can we discuss the benefits in the context of agents, e.g. reflection and retroformer, to make it more impactful? this property is very exciting.}

% compare with critique models
\begin{table*}[]
\vspace{-10pt}
\caption{Feedback quality of each model on different datasets rated by GPT-4. The best performance on each dataset is marked in \textbf{bold}, and the second has been \underline{underlined}.}
\centering
\resizebox{\textwidth}{!}{
\begin{tabular}{@{}l|ccccccccc|c@{}}
\toprule
\textbf{Model}   & \textbf{PIQA}        & \textbf{OBQA}        & \textbf{\begin{tabular}[c]{@{}c@{}}Common-\\ senseQA\end{tabular}} & \textbf{\begin{tabular}[c]{@{}c@{}}Alpaca-\\ Farm\end{tabular}} & \textbf{\begin{tabular}[c]{@{}c@{}}Fair-\\ Eval\end{tabular}} & \textbf{\begin{tabular}[c]{@{}c@{}}Human-\\ Eval\end{tabular}} & \textbf{MBPP}        & \textbf{MATH}        & \textbf{GSM8K}       & \textbf{Avg.} \\ \midrule
\texttt{gpt-3.5-turbo}    & \textbf{6.08}        & \textbf{6.12}        & \textbf{6.04}                                                       & \textbf{6.44}        & \textbf{6.32}        & \textbf{6.14}        & \textbf{6.48}        & \textbf{5.98}        & \textbf{5.94}        & \textbf{6.17}    \\
LLaMA2-13B-Chat & 5.92                 & 5.04                 & 5.66                                                                & 5.26                 & 5.74                 & 4.64                 & 4.82                 & 3.88                 & 4.30                 & 5.03             \\
Vicuna-13B-v1.5       & 5.66                 & 5.58                 & 5.42                                                                & 5.58                 & 5.82                 & 4.86                 & 5.20                 & 4.56                 & 4.84                 & 5.28             \\
WizardLM-13B-v1.2     & 5.90                 & 5.52                 & 5.82                                                                & 5.66                 & 5.88                 & 5.28                 & 5.34                 & 4.30                 & 4.90                 & 5.40             \\
\midrule
Shepherd-13B     & 3.48                 & 3.64                 & 3.48                                                                & 3.04                 & 3.30                 & 3.08                 & 3.20                 & 3.10                 & 2.76                 & 3.23             \\
SelFee-13B       & \underline{6.00}           & 5.32                 & 5.74                                                                & 5.88                 & 5.94                 & 4.84                 & 5.12                 & 4.46                 & 5.40                 & 5.41             \\
UltraCM-13B      & \underline{6.00}           & \textbf{6.12}        & \underline{6.02}                                                          & \underline{5.98}           & \underline{6.18}           & \underline{5.74}           & \underline{5.56}           & \underline{5.84}           & \underline{5.88}           & \underline{5.92}   \\ \bottomrule
\end{tabular}

}
\label{tab:feedback_quality}
\vspace{-15pt}
\end{table*}

\subsection{Capability Experiments}
\label{sec:app_capa}
We use nine datasets in Section~\ref{sec:capa} to test the model capability. For world knowledge, we adopt NaturalQuestions~\citep{kwiatkowski2019natural} and TriviaQA~\citep{joshi2017triviaqa}. For commonsense reasoning, we use PIQA~\citep{bisk2020piqa}, HellaSwag~\citep{zellers2019hellaswag}, OpenBookQA~\citep{mihaylov2018can}, and ARC~\citep{clark2018think}. For reading comprehension, we use BoolQ~\citep{clark2019boolq}, RACE~\citep{lai2017race} and MultiRC~\citep{khashabi2018looking}.

For evaluation, we simply ask models to answer the questions directly with answers (e.g. with options A, B, C, D or Yes/No). We then match the output with the ground truth and calculate the exact match scores. 

\subsection{Question Type Breakdown}
\label{sec:app_type}
Table~\ref{tab:ultra} reports the type-specific performance of our model and baselines compared with \texttt{gpt-3.5-turbo}. As is shown, our UltraLM-13B-PPO gets the highest average score, especially excels on the commonsense, world knowledge as well as professional knowledge questions. In the meantime, our model does not show advantages in math and reasoning tasks, which is consistent with the above results. 
\begin{table}[]
\caption{Relative scores (\%) versus \texttt{gpt-3.5-turbo} across different question types on UltraChat evaluation set. }
\resizebox{\textwidth}{!}{%
\begin{tabular}{@{}l|c|cc|cc|cc|cc|c|c@{}}
\toprule
\multicolumn{1}{c|}{\multirow{2}{*}{\textbf{Model}}} & \textbf{Vicuna} & \multicolumn{2}{|c|}{\textbf{Commonsense}} & \multicolumn{2}{c|}{\textbf{World Knowledge}} & \multicolumn{2}{c|}{\textbf{Professional Knowledge}} & \multicolumn{2}{c|}{\textbf{Ability}} & \multirow{2}{*}{\textbf{Writing}} & \multirow{2}{*}{\textbf{Overall}} \\
\multicolumn{1}{c|}{} & \textbf{Set} & \textbf{Easy} & \textbf{Moderate} & \textbf{Easy} & \textbf{Difficult} & \textbf{Physics} & \textbf{Biology} & \textbf{Math} & \textbf{Reasoning} &  &  \\ \midrule
\textbf{UltraLM-13B} & 95.6 & 113.7 & 106.8 & 111.7 & 103.3 & 102.1 & 105.1 & 89.7 & 71.0 & 98.6 & 98.8 \\
\textbf{Vicuna-13B-v1.3} & 93.2 & 113.4 & 106.4 & 109.6 & 107.1 & 106.0 & 108.9 & 84.7 & 79.0 & 98.4 & 98.8 \\
\textbf{Vicuna-13B-v1.5} & 95.7 & 115.8 & 106.6 & 104.9 & 105.0 & 100.1 & 101.2 & 94.8 & 73.2 & 99.1 & 99.0 \\
\textbf{LLaMA2-13B-Chat} & 97.1 & 114.6 & 108.5 & 109.3 & 107.7 & 105.9 & 108.0 & 91.3 & 75.0 & 98.6 & 100.2 \\
\textbf{Vicuna-33B-v1.3} & 98.5 & 113.4 & \textbf{114.0} & 105.1 & 109.0 & 109.9 & 112.8 & 84.4 & \textbf{86.7} & 103.0 & 102.4 \\
\textbf{WizardLM13B-v1.1} & 100.7 & 113.9 & 112.1 & 106.9 & 113.0 & 108.1 & 110.7 & 89.9 & 76.8 & 102.6 & 102.6 \\
\textbf{LLaMA2-70B-Chat} & 100.5 & 116.5 & 106.7 & 111.5 & 109.0 & 106.6 & 109.4 & \textbf{99.0} & 77.6 & 103.6 & 103.2 \\
\textbf{OpenChat-13B-v3.2super} & 98.6 & 121.2 & 112.6 & 116.1 & 110.1 & 106.0 & 110.0 & 89.3 & 82.9 & \textbf{104.7} & 103.9 \\
\textbf{WizardLM13B-v1.2} & \textbf{102.5} & 122.0 & 110.3 & 114.3 & 111.7 & 108.6 & 109.0 & 96.3 & 79.7 & 103.8 & 104.9 \\
\textbf{UltraLM-13B-PPO} & 97.7 & \textbf{123.5} & 113.6 & \textbf{131.1} & \textbf{118.4} & \textbf{113.2} & \textbf{120.2} & 93.0 & 78.8 & 101.7 & \textbf{105.7} \\ \bottomrule
\end{tabular}
}
\label{tab:ultra}
\end{table}

\section{Case Study}
\subsection{UltraFeedback}
\label{sec:case_uf}
We present a case sampled from ShareGPT in Table~\ref{tab:case}. The four assistants sampled to complete this instruction are Falcon-40B-Instruct, GPT-4, LLaMA2-70B-Chat, and UltraLM-13B. We also present GPT-4 annotations on the aspect of instruction following. GPT-4 accurately points out the lack of details in Falcon-40B-Instruct and GPT-4 responses.

\begin{table}[]
\small
\caption{An example of \textsc{UltraFeedback} comparing generations from 4 models from the aspect of instruction-following.}
\resizebox{\textwidth}{!}{%
\begin{tabular}{@{}p{12cm}p{5cm}@{}}
\toprule
\multicolumn{2}{l}{\textbf{User:} what's the general consensus best time to take vitamin D supplements.} \\ 
\midrule
\textbf{Response} & \begin{tabular}[c]{@{}p{5cm}@{}}\textbf{Annotation} \\\textbf{(Instruction-Following)} \end{tabular} \\
\midrule
\begin{tabular}[c]{@{}p{12cm}@{}} \textbf{\texttt{Falcon-40B-Instruct}}: A good rule of thumb is to take vitamin D supplements in the morning or afternoon so that they can be properly absorbed by the body. However, it is best to consult with a healthcare professional to determine a specific dosing regimen for your individual needs. \end{tabular} & \begin{tabular}[c]{@{}p{5cm}@{}} \textbf{Rating}: 3\\
\textbf{Rationale}: The text partially complies with the instruction, providing a general rule of thumb for taking vitamin D supplements. However, it lacks detail and does not address the various factors that may influence the best time to take the supplements. \end{tabular}\\
\midrule
\begin{tabular}[c]{@{}p{12cm}@{}}\textbf{\texttt{GPT-4}}: The general consensus on the best time to take vitamin D supplements varies, as individual needs and absorption rates can differ. However, many experts recommend taking vitamin D supplements with a meal containing healthy fats for better absorption.\\Some studies suggest that taking vitamin D supplements in the morning may be preferable to nighttime due to its potential effects on sleep. Keep in mind that it is crucial to follow the dosage recommended by a healthcare professional, as individual needs may vary. As always, it's best to consult with a healthcare provider for personalized advice on supplement timing and dosage.\end{tabular} & \begin{tabular}[c]{@{}p{5cm}@{}} \textbf{Rating}: 4\\
\textbf{Rationale}: The text almost fully aligns with the instruction, providing a general consensus on the best time to take vitamin D supplements and mentioning individual factors. However, it could provide more detail on the various factors that may influence the best time to take the supplements. \end{tabular}\\
\midrule
\begin{tabular}[c]{@{}p{12cm}@{}}\textbf{\texttt{LLaMA2-70B-Chat}}: Hello! I'm happy to help you with your question.\\The general consensus among health professionals is that the best time to take vitamin D supplements depends on various factors, including your personal schedule, sun exposure, and health status. Here are some general guidelines to consider:\\1. Morning: Taking vitamin D supplements in the morning can help you start your day with adequate levels of this essential nutrient. This may be particularly helpful if you have a busy schedule and may not have time to take them later in the day.\\2. After sun exposure: Vitamin D is produced in the skin when it's exposed to sunlight. Taking a vitamin D supplement after spending time outdoors can help boost your levels and ensure you're getting enough.\\3. With a meal: Vitamin D is a fat-soluble vitamin, which means it's best absorbed when taken with a meal that contains healthy fats. Having a meal with a source of fat, such as avocado, nuts, or seeds, can enhance the absorption of vitamin D.\\4. Consistency: The most important thing is to establish a consistent routine and take your vitamin D supplements at the same time every day. This will help maintain a stable level of vitamin D in your body.\\It's important to note that excessive intake of vitamin D can be harmful, so it's essential to consult with a healthcare professional to determine the appropriate dosage for your individual needs. They can help you identify the best time to take vitamin D supplements based on your specific circumstances.\\Remember, it's always a good idea to consult with a healthcare professional before starting any new supplement regimen. They can assess your individual needs and provide personalized recommendations for your health and well-being.\end{tabular} & \begin{tabular}[c]{@{}p{5cm}@{}} \textbf{Rating}: 5\\
\textbf{Rationale}: The text comprehensively addresses the task goal by providing a general consensus on the best time to take vitamin D supplements, considering various factors. It also acknowledges the importance of consulting with a healthcare professional for personalized advice. \end{tabular}\\
\midrule
\begin{tabular}[c]{@{}p{12cm}@{}}\textbf{\texttt{UltraLM-13B}}: According to current research and guidelines, the general consensus is that the best time to take vitamin D supplements is either in the morning or early afternoon, as our bodies produce more vitamin D when exposed to sunlight during this time. However, it's important to note that the optimal time to take vitamin D supplements can vary depending on individual factors such as sun exposure, geographical location, and overall health.\\For people living in areas with limited sunlight or those who spend most of their time indoors, taking vitamin D supplements with a meal can be an effective way to ensure adequate intake. Some research also suggests that taking larger doses of vitamin D, such as 2,000-5,000 IU (international units), as a single dose a week or every few weeks, may be more effective for increasing vitamin D levels and maintaining them over time.\\Keep in mind that the optimal time to take vitamin D supplements may also be influenced by medications or supplements you are currently taking, so it's best to consult with a healthcare professional for personalized advice based on your individual needs and circumstances. \end{tabular} & \begin{tabular}[c]{@{}p{5cm}@{}} \textbf{Rating}: 5\\\textbf{Rationale}: The text fully aligns with the instruction, providing a general consensus on the best time to take vitamin D supplements and considering individual factors. It also emphasizes the importance of consulting with a healthcare professional for personalized advice. \end{tabular}\\
\bottomrule
\end{tabular}
}
\label{tab:case}
\end{table}

\subsection{Best-of-$n$ Sampling}
\label{sec:case_bon}
We present a case comparing best-of-$1$, best-of-$2$, and best-of-$16$ sampling in Table~\ref{tab:case_bo16}. 
The best-of-$1$ sample is incorrect at all, so it received a low reward of $-0.73$. To compare, the best-of-$2$ sample is partially correct, as it only addressed the @ symbol in decorating \texttt{staticmethod}. The example it provided is also problematic. 
The best-of-$16$ sample, however, correctly illustrated the usage of the @ symbol in Python with a valid example. Thus it received a much higher reward from the reward model. Another worth-noting point is that the best-of-$2$ sample is longer than the best-of-$16$ sample, which verified that our UltraRM does not depend on length shortcut to evaluate the response quality.
\begin{table}[]
\small
\caption{Comparative cases of best-of-$1$, best-of-$2$, and best-of-$16$ sampling. }
\begin{tabular}{@{}p{\textwidth}@{}}
\toprule
\textbf{User:} what does the @ in python do \\ 
\midrule
\begin{tabular}{@{}p{\textwidth}@{}}
\textbf{Best-of-$1$}: In Python, the @ symbol is used as a way to reference a class within its own namespace. This is often referred to as a class object. When you create an instance of a class, you are creating a new object that is stored in the namespace of the class. Using the @ symbol before a class declaration allows you to bypass. {\color{violet}\textbf{Reward: $-0.73$}}
\end{tabular}\\
\midrule
\begin{tabular}{@{}p{\textwidth}@{}}
\textbf{Best-of-$2$}: 
The @ symbol in Python is used in front of a class to indicate that the class is decorated with the \texttt{@staticmethod} decorator. This decorator is sandwiched between the class definition and the name of the class. A static method in Python is a method that can be called directly on a class, like a regular method, but does not have access to the instance data (self). It usually takes no parameters and does not modify the object's state directly. Instead, it operates on the class itself. As a result, if you define a method as \texttt{staticmethod}, it will become a static method and you can call it on an instance of the class without having to create an instance of the class first. For example, a \texttt{Rectangle} class could define the \texttt{perimeter} method as a static method:
\begin{python}[basicstyle=\small]
@staticmethod
def perimeter(width, height):
    return 2*width + 2*height
\end{python}
You can then call this \texttt{perimeter} method on a \texttt{Rectangle} instance as follows:
\begin{python}[basicstyle=\small]
rect = Rectangle(10, 5)
print(rect.perimeter()) # Output: 40
\end{python}
In this example, the \texttt{Rectangle} class defines a \texttt{perimeter} method, which is static by nature. The \texttt{@staticmethod} decorator simply serves to inform the Python interpreter that the method you are defining is indeed intended to be a static method. Moreover, you could also use this decorator with \texttt{classmethod} or \texttt{lamdba} function syntax. The `@staticmethod` decorator is by default aligned with \texttt{classmethod}, but the other formats are used for cases where you need more control over the function's self context, or more specific function memory. In summary, the @ symbol is a Python decorator that converts a method into a static method, which is able to be called on an instance of the class without requiring the creation of an actual instance. This functionality makes your code cleaner, more readable, and easier to maintain. Please note that Python has several other symbols and decorators that can modify the behavior of your code or enrich the functionality of your objects, such as \texttt{@staticmethod}, \texttt{@classmethod}, \texttt{@property}, and \texttt{@newmethod}. However, these will generally require some level of understanding and configuration before they can be effectively adapted for your applications. Remember that these decorators should not be confused with the general syntax to create a static method. If you don't see the @ symbol in front of a class method, then it is by default considered as a method. Adding static to the method implies decorating the method with the \texttt{staticmethod} decorator. To keep the code interesting, decorators are an elegant and expressive way to integrate various functional programming concepts into Python, improving encapsulation and abstraction across scripts and classes.
{\color{violet}\textbf{Reward: $-0.10$}}
\end{tabular} \\
\midrule
\begin{tabular}{@{}p{\textwidth}@{}}
\textbf{Best-of-$16$}: 
In Python, the @ symbol is used to define a decorator, which is a way to add functionality to or modify the behavior of functions or objects (not class methods). A decorator is essentially a decoration around the original function, executing some arbitrary code or modifying the function's properties before the function is executed.
When a function is correctly decorated using the @ symbol, a decorator function is applied to it. The decorator function can then modify the behavior of the original function, by, for example, adding arguments, modifying attributes of the function, or executing additional code before or after the function is executed.
Here's an example of how to use a simple decorator to add some additional functionality to a function:\\
\begin{python}[basicstyle=\tiny]
from functools import wraps
def login_required(func):
    @wraps(func)
    def wrapper(username, *args, **kwargs):
        if username != "secret":
            raise Exception("Please enter the correct username and password.")
        return func(username, *args, **kwargs)
    return wrapper
@login_required
def secret_page(username):
    print(f"Hello, {username}! You are now on the secret page.")
\end{python}
In this example, we are defining a decorator \texttt{login\_required} to handle the authentication of our user. It will raise an exception if the provided `username` argument is incorrect. And then we use this decorator to wrapper the interesting function \texttt{secret\_page} with this restriction.
This approach allows you to modify the behavior of a function at runtime, making it a very flexible and powerful tool for Python development. {\color{violet}\textbf{Reward: $0.42$}}
\end{tabular}
\\
\bottomrule
\end{tabular}
\label{tab:case_bo16}
\end{table}

\subsection{UltraLM-13B-PPO}
\label{sec:case_ppo}
We compare UltraLM-13B before and after PPO from an AlpacaEval case in Table~\ref{tab:case_ppo}. As we can see, the original UltraLM-13B made mistakes in describing Lady Gaga's characters in American Horror Story. The UltraLM-13B-PPO, on the other hand, correctly answered the question.

\begin{table}[]
\small
\caption{Comparative cases of UltraLM-13B and UltraLM-13B-PPO. }
\begin{tabular}{@{}p{\textwidth}@{}}
\toprule
\textbf{User:} who does lady gaga play in american horror story \\ 
\midrule
\begin{tabular}{@{}p{\textwidth}@{}}
\textbf{UltraLM-13B}: The question is asking for the character that Lady Gaga has played in the television series American Horror Story. Lady Gaga has appeared in multiple seasons of the show and has played different characters in each one. The following are the characters Lady Gaga has played in the series: \\ 1. Season 1: American Horror Story (2011) - Gaga played the role of The Countess or Elizabeth Johnson, a wealthy and sophisticated owner of a Los Angeles apartment building who resides with her ventriloquist dummy. \\ 2. Season 6: American Horror Story: Hotel (2016) - In this season, Lady Gaga portrayed German singer-songwriter Elizabeth along with her twin brother Dennis. They have gained fame in the 1970s with their horror rock collaboration, The Countess. \\ Gaga's portrayal in American Horror Story has added a unique dimension to the series, showcasing her acting talent, and has become a fan-favorite character.
\end{tabular}\\
\midrule
\begin{tabular}{@{}p{\textwidth}@{}}
\textbf{UltraLM-13B-PPO}: Lady Gaga played the role of The Countess in American Horror Story: Hotel. The Countess is a mysterious and enigmatic character who runs the Hotel Cortez, a luxurious and decadent hotel that serves as a front for a series of gruesome murders. The Countess is a seductive and charismatic figure who preys on her victims, using her beauty and charm to lure them into her web. Throughout the season, viewers are taken on a twisted and terrifying journey as they uncover the dark secrets of the Hotel Cortez and the sinister forces that lurk within its walls. Lady Gaga's performance in American Horror Story: Hotel was widely praised for its depth, complexity, and sheer intensity, cementing her status as one of the most talented and versatile actresses of her generation.
\end{tabular}
\\
\bottomrule
\end{tabular}
\label{tab:case_ppo}
\end{table}

\subsection{UltraCM-13B}
To comprehensively evaluate and analyze the critiques our model generates, we conducted case study in Table~\ref{tab:case_ultracm}. On GSM8K benchmark, we find that UltraCM-13B could give meaningful critiques on incorrect answers, pointing out flaws and provide suggestions. However, the suggestions are sometimes unnecessarily long, which indicates the imperfection of the model. We will continue to improve the critique quality in future work.
\begin{table}[]
\small
\caption{Case of UltraCM-13B on GSM8K.}
\begin{tabular}{@{}p{\textwidth}@{}}
\toprule
\textbf{User:} A hotel in the Philippines charges 1000 pesos for a 12-hour stay or 1600 pesos for a 24-hour stay. After 12 hours, visitors have the option to add 70 pesos for every additional hour. Cameron arrives at 5 pm at the hotel and wants to leave at 10 am the next morning. He decides to go with the option of adding on 70 pesos for every hour after the 12-hour mark instead of paying for 24 hours. How much money would Cameron save? {\color{violet}\textbf{Answer: $250$}} \\ 
\midrule
\begin{tabular}{@{}p{\textwidth}@{}}
\textbf{Vicuna-33B}: First, let's determine the cost of Cameron's stay if he were to stay for 12 hours. Since he arrives at 5 pm, his stay would actually be 13 hours (from 5 pm to 6 am). The cost for a 12-hour stay is 1000 pesos.\\Next, let's calculate the cost of a 24-hour stay. The cost for a 24-hour stay is 1600 pesos.\\Now, we need to calculate the cost of Cameron's stay if he were to stay for 13 hours and pay the additional 70 pesos for every hour after the 12-hour mark. For the first 12 hours, the cost is 1000 pesos. For the additional 1-hour period (from 12 hours to 13 hours), the cost is 70 pesos. Therefore, the total cost for Cameron's 13-hour stay would be 1000 pesos + 70 pesos = 1070 pesos.\\Now, we can calculate the amount Cameron would save by choosing the 13-hour option over the 24-hour option. The savings would be the difference between the two costs:\\ Savings = 1600 pesos (24-hour cost) - 1070 pesos (13-hour cost)\\ avings = 530 pesos\\ So, Cameron would save 530 pesos by choosing the option of adding 70 pesos for every additional hour after the 12-hour mark instead of paying for 24 hours.
\end{tabular}\\
\midrule
\begin{tabular}{@{}p{\textwidth}@{}}
\textbf{UltraCM-13B Critique}: Your calculation of the hotel stay cost is incorrect. Cameron's stay from 5 pm to 10 am is 17 hours, not 13 hours. You should calculate the cost for a 17-hour stay instead. Also, you didn't need to calculate the savings by staying for 13 hours. Just focus on the correct calculation for the 17-hour stay. The rest of your answer is detailed and understandable, but the numbers are wrong due to the incorrect calculation of the hotel stay hours. Please revise this. \\ For improvement, always double-check your calculations to avoid such mistakes. Make sure to understand the problem clearly before you start solving it. Focus on what the question is asking you to find and ignore unnecessary details that might confuse you. As an AI, accuracy is key. A small mistake can lead to a wrong conclusion and misinformation, which is not our goal as AI assistants. Keep practicing and focus on your accuracy, try to avoid rush decisions and take your time to understand the questions fully before answering them. \\ Overall Score: 6
\end{tabular}
\\
\bottomrule
\end{tabular}
\label{tab:case_ultracm}
\end{table}

\section{Prompt}
\subsection{Principle}
\label{sec:app_prin}
In model completion, we sample one system prompt to align model behavior from different aspects. The human-written examples for each aspect are as follows:

\begin{figure*}[!th]
\begin{tcolorbox}[
    colback=white, % background color
    colframe=gray, % frame color
    % coltitle=black, % title color
    title=\textbf{Human-written Principle Prompts}, % box title
    fonttitle=\bfseries\large, % title font
    arc=4mm, % corner radius
]
%\begin{markdown}

**Honesty:**

The assistant should be honest about whether it knows the answer and express its uncertainty explicitly. Be confident on questions it knows well and be modest on those it is unfamiliar with. Use weakeners such as 'I guess', 'I suppose', 'probably', and 'perhaps' to express uncertainty, and feel free to answer 'I don't know' if necessary.

---

**Verbalized Calibration:**

The assistant should express its confidence as a scalar at the end of the response. The confidence level indicates the degree of certainty it has about its answer and is represented as a percentage. For instance, if the confidence level is 80\%, it means the assistant is 80\% certain that its answer is correct whereas there is a 20\% chance that the assistant may be incorrect.

The format is as follows:
[Question]
[Answer]
Confidence: [The assistant's confidence level, numerical numbers only, e.g. 80\%]
Here, tags like [Question] and [Answer] are placeholders and should be omitted in the response.

---

**Truthfulness:**

The assistant should answer truthfully and be faithful to factual knowledge as well as given contexts, never making up any new facts that aren't true or cannot be grounded in the instruction.

---

**Helpfulness:**

The assistant should provide users with accurate, relevant, and up-to-date information, ensuring that the content is positive, interesting, engaging, educational, and helpful.

%\end{markdown}
\end{tcolorbox}
\end{figure*}

\subsection{Annotation}
\label{sec:app_anno}
% \subsubsection{Preference Feedback}
We first showcase the template that prompts GPT-4 to annotate the quality of four given completions from the aspect of instruction following. Then, we present the template to annotate critique feedback.

\begin{figure*}[!th]
\begin{tcolorbox}[
    colback=white, % background color
    colframe=gray, % frame color
    %coltitle=black, % title color
    title=\textbf{Annotation Template for Instruction Following}, % box title
    fonttitle=\bfseries\large, % title font
    arc=4mm, % corner radius
]
%\begin{markdown}

**Instruction Following Assessment**

Evaluate alignment between output and intent. Assess understanding of task goals and restrictions.

**Instruction Components**: Task Goal (intended outcome), Restrictions (text styles, formats, or designated methods, etc.).

**Scoring**: Rate outputs 1 to 5:

1. **Irrelevant**: No alignment.

2. **Partial Focus**: Addresses one aspect poorly.

3. **Partial Compliance**:

    - (1) Meets goals or restrictions, neglecting others.
    
    - (2) Acknowledges both but slight deviations.
    
4. **Almost There**: Near alignment, minor deviations.

5. **Comprehensive Compliance**: Fully aligns, meets all requirements.
%\end{markdown}
\end{tcolorbox}
\end{figure*}

\begin{figure*}[!th]
\begin{tcolorbox}[
    colback=white, % background color
    colframe=gray, % frame color
    % coltitle=black, % title color
    title=\textbf{Annotation Template for Critique Feedback}, % box title
    fonttitle=\bfseries\large, % title font
    arc=4mm, % corner radius
]
%\begin{markdown}
Given my answer to an instruction, your role is to provide specific and constructive feedback for me. You should find the best way for me to learn from your feedback and improve my performance. 

You should consider multiple aspects of my answer, including helpfulness, truthfulness, honesty, and to what extent the answer follows instructions.

---

\#\#\# Instruction

\{instruction\}

\#\#\# Answer

{completion}

---

Please act as a teacher and provide specific and constructive feedback. Besides describing the weaknesses of the answer, you should also provide specific suggestions to guide me toward understanding how to improve. Please note, however, that your suggestions should help me better complete the instructions, but you should not introduce new requirements that are not mentioned in the instructions. Your feedback should focus on enhancing my ability to think critically and respond accurately. However, never explicitly provide the reference answer, nor do polite phrases be required. Only respond with concise feedback in chat style. Finally, score the overall quality of the answer from 1 to 10, where 1 is the worst and 10 is the best.

\*Format\*

\#\#\# Feedback

[Your feedback]

Overall Score: [1-10]

---

\#\#\# Feedback
%\end{markdown}
\end{tcolorbox}
\end{figure*}

\end{document}